\documentclass[journal]{IEEEtran}
\ifCLASSINFOpdf
\else
\fi
\usepackage[breaklinks,colorlinks]{hyperref}
\usepackage{dsfont}
\usepackage{subfig}
\usepackage{graphicx}
\usepackage{amsmath}
\usepackage{amssymb}
\usepackage{orcidlink}
\usepackage{booktabs}
\usepackage{amsmath, amssymb}
\usepackage{makecell}
\usepackage{multirow}
\usepackage{gensymb}
\usepackage{bbding}
\usepackage{float}
\usepackage{tcolorbox}
\usepackage{cleveref}
\usepackage{algpseudocode}
\usepackage{algorithm}

\begin{document}

\title{Towards Top-Down Reasoning: An Explainable Multi-Agent Approach for Visual Question Answering}

\author{Zeqing~Wang~\orcidlink{0009-0006-6389-6678},
        Wentao~Wan,
        Qiqing~Lao,
        Runmeng~Chen,
        Minjie~Lang,
        Xiao~Wang,
        Keze~Wang~\orcidlink{0000-0002-7817-8306},
        Liang~Lin~\orcidlink{0000-0003-2248-3755}, \IEEEmembership{Fellow, IEEE}

\thanks{Zeqing Wang, Wentao Wan, Qiqin Lao Keze Wang, and Liang Lin are with the School of computer science and Engineering, Sun Yat-sen University, Guangzhou, Guangdong 510000, China. Runmeng Chen is with the South China Normal University, Guangzhou, Guangdong 510000, China. Minjie Lang is with the Northeastern University, Shenyang, Liaoning 110000, China. Xiao wang is with the Anhui University, Hefei, Anhui 230601, China (E-mail: wangzq73@mail2.sysu.edu.cn; kezewang@gmail.com; lianglin@ieee.org).}

}

\markboth{Submitted to IEEE Transactions on Multimedia}%
{Shell \MakeLowercase{\textit{et al.}}: Bare Demo of IEEEtran.cls for IEEE Journals}

\maketitle

\begin{abstract}
Recently, to comprehensively improve Vision Language Models (VLMs) for Visual Question Answering (VQA), several methods have been proposed to further reinforce the inference capabilities of VLMs to independently tackle VQA tasks rather than some methods that only utilize VLMs as aids to Large Language Models (LLMs). However, these methods ignore the rich common-sense knowledge inside the given VQA image sampled from the real world. Thus, they cannot fully use the powerful VLM for the given VQA question to achieve optimal performance. Attempt to overcome this limitation and inspired by the human top-down reasoning process, i.e., systematically exploring relevant issues to derive a comprehensive answer, this work introduces a novel, explainable multi-agent collaboration framework by leveraging the expansive knowledge of Large Language Models (LLMs) to enhance the capabilities of VLMs themselves. Specifically, our framework comprises three agents, i.e., \textit{Responder}, \textit{Seeker}, and \textit{Integrator}, to collaboratively answer the given VQA question by seeking its relevant issues and generating the final answer in such a top-down reasoning process. The VLM-based \textit{Responder} agent generates the answer candidates for the question and responds to other relevant issues. The \textit{Seeker} agent, primarily based on LLM, identifies relevant issues related to the question to inform the \textit{Responder} agent and constructs a Multi-View Knowledge Base (MVKB) for the given visual scene by leveraging the build-in world knowledge of LLM. The \textit{Integrator} agent combines knowledge from the \textit{Seeker} agent and the \textit{Responder} agent to produce the final VQA answer. Extensive and comprehensive evaluations on diverse VQA datasets with a variety of VLMs demonstrate the superior performance and interpretability of our framework over the baseline method, e.g., 5.7\% improvement on VQA-RAD and 5.2\% on Winoground in the zero-shot setting without extra training cost.
\end{abstract}


\IEEEpeerreviewmaketitle

\section{Introduction}
\label{sec:intro}
\begin{figure}[!t]
  \centering
  \includegraphics[width=\linewidth]{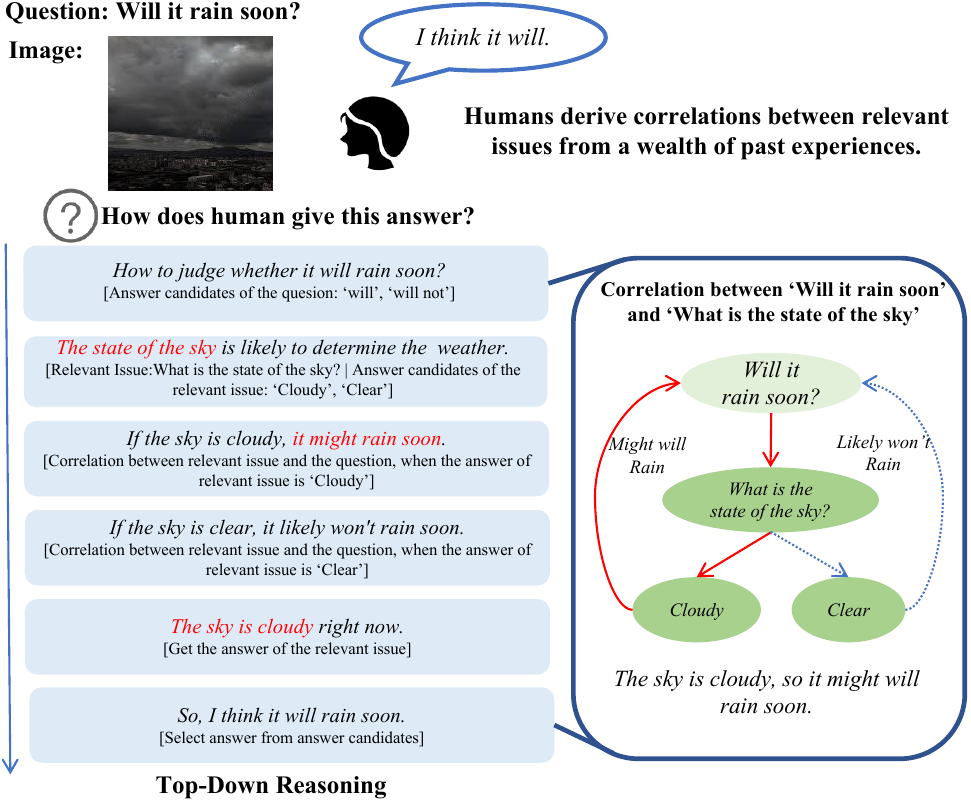}
  \caption{The illustration of how humans solve VQA tasks in a top-down reasoning process. When solving a question, humans find relevant issues that can distinguish between the answer candidates. The different answers to the relevant issue make humans select a different answer candidate based on the correlation between the two.
  For example, cloudy skies mean that it will rain, while clear skies do not. This correlation is summarized by humans through a wealth of world knowledge.}
    \label{fig:human_topdown}
\end{figure}

\IEEEPARstart{W}{hat} would you do if asked whether it would rain soon? Before answering, you might probably look up at the sky to check its state. If it is cloudy, you might believe that it will rain soon, as shown in Figure \ref{fig:human_topdown}. This is a process of top-down reasoning by humans. In the realm of Visual Question Answering (VQA)~\cite{vqa, vqav2}, humans are adept at leveraging such a top-down reasoning process to address questions by synthesizing relevant issues, thereby facilitating the resolution of complex questions even if they are not sure about their answer. This capability is underpinned by an intricate web of correlations among diverse questions, distilled from a wealth of past experiences, such as the heuristic that a cloudy sky may presage rainfall.

In contrast, contemporary approaches for VQA predominantly conceptualize the task as an end-to-end process~\cite{nguyen2022coarse, Goyal_Khot_Agrawal_Summers-Stay_Batra_Parikh_2019,yang2021justask, yuan2023self,jin2021ruart,mao2022positional}. This methodology exhibits palpable limitations in terms of consistency and generalizability across varied datasets~\cite{chen2023measuring, selvaraju2020squinting}. 
To address the limitations of single-step response mechanisms in VQA tasks, some methods~\cite{idealgpt,lan:mm2023} have been proposed by treating Vision Language models (VLMs) as a limited tool to assist other models, e.g., Large Language Models (LLMs) in a step-by-step fashion. However, due to that these methods model the reasoning process along with the LLM rather than VLM, they can not fully release the potential generalization capability of VLMs under various scenarios~\cite{XING2024103885}.

However, directly modeling the reasoning process along with the VLM is quite challenging, due to the difficulty of implicitly using the relevant knowledge of the VLM for the given VQA question to provide the optimal answer.  
Some works~\cite{khan2024exploring,wang2024q} proposed to consider the sub-questions as visual clues that might be beneficial for the reasoning process of VLMs. However, since the existing VLMs are trained on limited vision-language data~\cite{10448321,song2024missing}, these works cannot ensure these visual clues are leveraged by VLMs in such an appropriate manner. Hence, their performance gains are somewhat incremental~\cite{zhang2024visualquestiondecompositionmultimodal}.



\begin{figure}[!t]
  \centering
  \includegraphics[width= 0.65 \columnwidth]{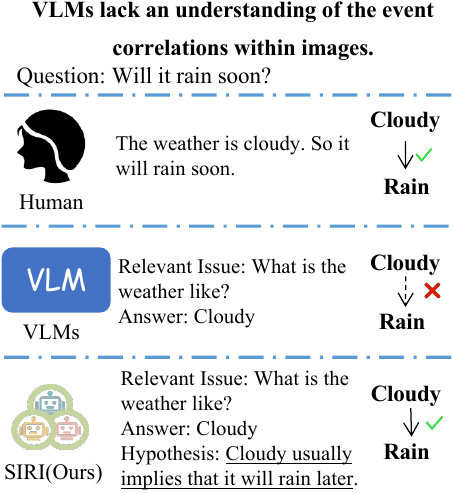}
    \caption{Demonstration of how the top-down reasoning process helps VLM answer questions more accurately. Although VLM's powerful capabilities can handle a variety of arbitrary question types, it lacks the correlation between different issues. Our SIRI enhances VLM's ability by injecting this type of connectivity information into VLM via hypothesis.}
    \label{fig:help_vlm}
\end{figure}



Attempt to overcome this limitation and inspired by the human top-down reasoning process (see Figure~\ref{fig:human_topdown}) that fully utilizes the common-sense knowledge inside the image, we propose the following intuitive assumption:
\begin{center}
\textit{\textbf{The performance of VLMs can be enhanced through the emulation of the human top-down reasoning process.}}
\end{center}

This delineates the core objective of our work, as shown in Figure~\ref{fig:help_vlm}, emulating the human top-down reasoning process and enhancing the performance of VLMs in VQA tasks. 

However, enabling the top-down reasoning process in VQA tasks for VLMs is quite challenging. The ability to assume correlations among disparate events inherently demands a profound comprehension of the real world. Thanks to the rapid development of LLMs trained from huge data, synergizing the capabilities of LLMs with VLMs in the context of VQA tasks might be sufficient to provide complex relevance in the real world.
However, it is still challenging, primarily due to the significant linguistic and domain discrepancies between LLMs and VLMs~\cite{guo2023images,khan2023q} and the complexity between multimodalities~\cite{vosoughi2024cross,wang2015learning}.




\begin{table}[!b]
\small
  \centering
      \caption{Comparison of training-free \textbf{(T-F)}, Component Oriented \textbf{(C-O)}, Interpretability \textbf{(I)} and Continuous Improvement Ability \textbf{(C-I)}
with other methods.}
  \resizebox{1\columnwidth}{!}{ 
    \begin{tabular}{cccccc}
    \toprule
    Type  & Method & T-F   & C-O   & I     & C-I \\
    \midrule
     \multirow{5}[2]{*}{LLM-Based} & IdeaGPT~\cite{idealgpt} & \checkmark & \XSolidBrush & \checkmark & \checkmark \\
           & MM-ReAct~\cite{yang2023mm} & \checkmark & \XSolidBrush & \checkmark & \checkmark \\
           & ViperGPT~\cite{vipergpt} & \checkmark & \XSolidBrush & \checkmark & \checkmark \\
           & Visprog~\cite{visualprogramming} & \checkmark & \XSolidBrush & \checkmark & \checkmark \\
           & CLOVA~\cite{gao2024clova} & \XSolidBrush & \checkmark & \checkmark & \checkmark \\
    \midrule
    \multirow{3}[2]{*}{\textbf{VLM-Based}} & Vanile & \checkmark & \checkmark & \XSolidBrush & \XSolidBrush \\
          & Q-D~\cite{khan2024exploring}   & \checkmark & \checkmark & \XSolidBrush & \XSolidBrush \\
          & \textbf{SIRI(Ours)} & \checkmark & \checkmark & \checkmark & \checkmark \\
    \bottomrule
    \end{tabular}%
    }

  \label{tab:compare_methods}%
\end{table}%

Inspired by the recent multi-agent methodologies~\cite{debate,li2024agents,mindstorms,lan2023collaborative}, we present a novel multi-agent collaboration framework, including three agents, i.e.,\textit{\textbf{S}eeker} for \textbf{I}ssues, \textit{\textbf{R}esponder}, and \textit{\textbf{I}ntegrator} \textbf{\textit{(SIRI)}}. 
Our SIRI is to emulate the top-down reasoning process inherent in human cognition and reinforce VLMs' performance on VQA tasks via a collaborative relationship among these three agents. Specifically, the \textit{Responder} agent, based on VLM, generates answer candidates based on the input question-image pair, serving as an initial step in the top-down reasoning process. The \textit{Seeker} agent, based on LLM, capitalizes on both the question and its answer candidates to generate a set of relevant issues that exhibit a relationship with the question. Subsequently, the \textit{Seeker} agent employs the \textit{Responder} agent to obtain answer candidates for the relevant issues. By combining the answer candidates derived from the question with the responses of the relevant issues, the \textit{Seeker} agent formulates hypotheses (such as `if the sky is cloudy, it will rain', etc.). Then, leveraging the powerful understanding ability of LLM, the \textit{Seeker} agent can obtain a confidence score for this hypothesis (`if the sky is cloudy, it will rain. 0.8', etc), which means the confidence of the hypothesis occurring in the world, ultimately achieving modeling of the relationship between the question and the issue. This step aims to mimic the human top-down reasoning process of searching for relevant issues to solve the VQA task.
Finally, the \textit{Integrator} agent generates the answer to the question with the information given by the \textit{Seeker} agent, as the last step in the top-down reasoning process, i.e., providing the final answer with the help of hypotheses and their confidence score. 


As shown in Table~\ref{tab:compare_methods}, we provide a summary comparison of our SIRI against others from five aspects: whether it is based on LLM or VLM (\textbf{LLM-Based or VLM-Based}), whether training-free (\textbf{T-F}), whether it targets a single VLM component-oriented (\textbf{C-O}) thus offering better generalizability, whether it is interpretable (\textbf{I}), and whether it can be continuous improvement (\textbf{C-I}). Table~\ref{tab:compare_methods} accurately depicts the focal efforts of this paper's research.

In summary, the \textbf{main contributions} of our work are listed as follows: i) We propose a novel multi-agent collaboration framework, i.e., SIRI. By emulating the top-down reasoning process of humans, our SIRI can enhance the performance of VLMs on VQA tasks via external common-sense knowledge inside the image; 
ii) We introduce the assistive strategy, i.e., asking relevant issues and formulating hypotheses via LLM, to effectively mine event-related semantic information of the given questions for the VLM; iii) By leveraging the correlation between the relevant issues and the question, we construct a tailored Multi-View Knowledge Base for the given visual scene to explicitly perform confidence scores, which enables our SIRI framework to be consistent with the top-down human reasoning process and provides sufficient interpretations. 

The remainder is organized as follows. Related work is introduced in Section \ref{section:related}. The proposed multi-agent framework is described in detail in Section \ref{section:method}. The experiments are presented in Section \ref{section:exper}. The limitations and detailed research motivation of our SIRI are discussed in Section \ref{section:disscussion}. The conclusion is given in Section \ref{section:conclusion}.

\section{Related Work}\label{section:related}

\textbf{Visual Question Answering (VQA):} VQA is a traditional yet still highly challenging cross-modal task. Classification-based VQA models~\cite{vqa,nguyen2022coarse,jiang21x_ggm,han21focal} typically employ classification methods to solve tasks for specific datasets. With the development of multimodal models, an increasing number of works have adopted prompt-based zero-shot methods for VQA tasks~\cite{yang2021justask,lin2023video}, significantly expanding the range of VQA tasks a single model can handle. Some approaches address Knowledge-Based VQA tasks requiring external knowledge~\cite{guo2022_knowledge,conceptbert, outofthebox,krisp,li22ai,wu2023resolving} by integrating knowledge bases~\cite{wikidata,conceptbert,aser} to assist the model. Similarly, with the development of LLMs~\cite{llama,llama2,gpt3}, some methods use LLMs to assist~\cite{idealgpt,assistgpt,woodpecker} with or directly~\cite{PICa,promptcap,lan:mm2023} perform VQA tasks. Recently, some works~\cite{khan2024exploring,wang2024q} focus on question decomposition in VQA tasks, which aims to add more information derived from the sub-questions of the question for VLMs.
  
\textbf{Vision-Language Models (VLMs):} VLMs have demonstrated remarkable capabilities in various multi-modal tasks~\cite{flamingo, blip, blip2, bliva,llava,hong2023cogagent,wang2023cogvlm}. These models typically undergo a pretraining process on large-scale image-text datasets~\cite{schuhmann2022laion,ng2020understanding}. Subsequently, to adapt to a specific downstream task, they are fine-tuned using datasets relevant to the task~\cite{xiao2023clip,li2023llavamed,zhu2023not}. Leveraging their extensive pretraining datasets, VLMs exhibit strong performance across different multi-modal tasks, with several achieving state-of-the-art results in zero-shot settings. Nevertheless, the increasing size of these models presents a significant challenge in terms of the computational cost of pretraining~\cite{dai2023instructblip, bai2023qwen}. However, to effectively address complex real-world tasks, VLMs are required to possess robust common sense and knowledge~\cite{rao2023retrieval,chen2022unsupervised}. The high cost of pretraining poses a barrier to the development of more powerful VLMs. 

\textbf{LLMs for Visual Tasks:} LLMs have achieved a profound impact on the field of artificial intelligence~\cite{llama,llama2,gpt3}. Several approaches have leveraged LLMs as controllers for visual models in visual programming. Examples include Visual Programming~\cite{visualprogramming}, ViperGPT~\cite{vipergpt}, and AssistGPT~\cite{assistgpt}, which utilize LLMs in combination with Python code to address various visual tasks in a flexible method. Additionally, some works employ LLMs as knowledge bases to tackle the VQA task such as PICa~\cite{PICa} and PromptCap~\cite{promptcap} etc.  Similarly, IdealGPT~\cite{idealgpt} employs LLMs as questioners, generating sub-questions that facilitate reasoning about the main question. Woodpecker~\cite{woodpecker}, on the other hand, employs LLMs to evaluate the output of VLMs and rectify any inaccuracies in their responses. However, these methods delegate the final answer to VQA tasks to LLMs, which limits their application scenarios and overlooks the potential inherent in VLMs themselves.

\textbf{Multi-Agent Frameworks:} Numerous studies~\cite{agent_survey,li2024agents} have leveraged the diverse capabilities offered by LLMs, resulting in a significant body of work that centers around LLM-based agents. Within this context, certain approaches~\cite{debate,agent_self_col} have employed the technique of agent-based discussions to enhance problem-solving capabilities. Additionally, several studies~\cite{generative_agent,mindstorms} have drawn inspiration from sociological phenomena to construct multi-agent frameworks capable of emulating human behavior. Additionally, numerous works apply the multi-agent framework in the visual domain~\cite{yu2022learning,li22ai,lan2023collaborative}. 

\section{Proposed Method} \label{section:method}

\begin{figure*}[h]
  \centering
  \includegraphics[width=0.95\textwidth]{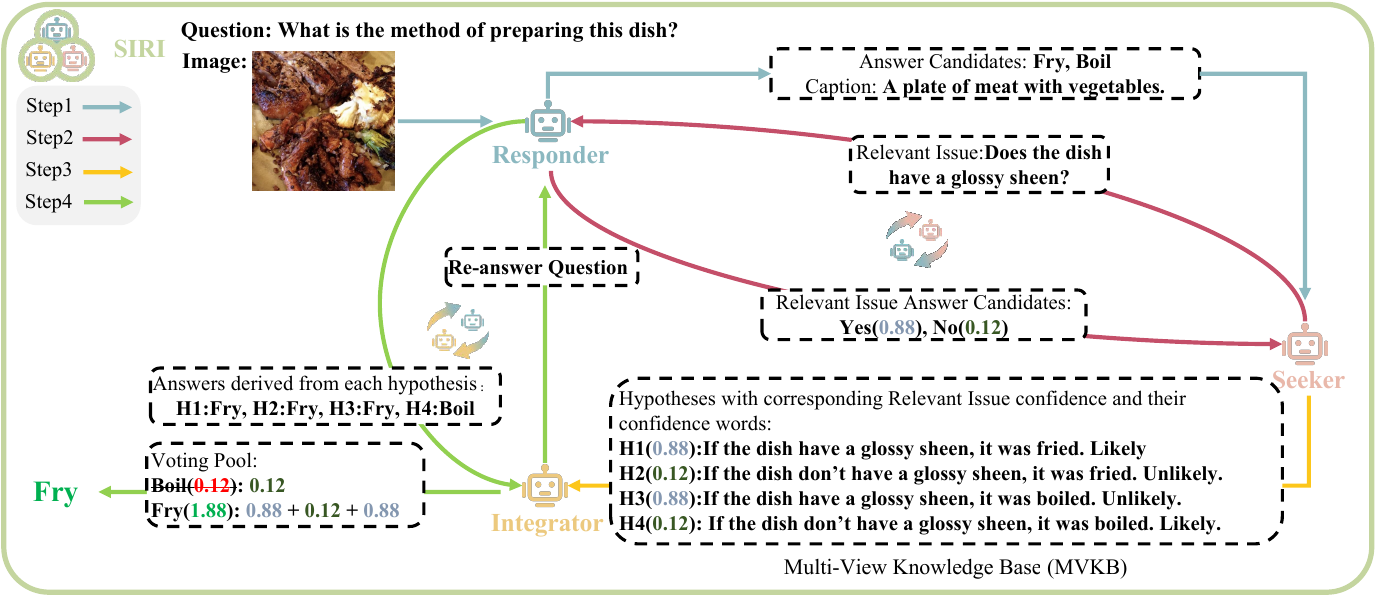}
  \caption{The interaction among agents in our SIRI. The \textit{Responder} agent receives a question-image pair and generates answer candidates for the given question. Additionally, the \textit{Responder} agent can also serve as a captioner, providing descriptive content for the image. The \textit{Seeker} agent leverages the answer candidates, the caption of the image and the capabilities of LLMs to generate relevant issues. With assistance from the \textit{Responder} agent, the \textit{Seeker} agent obtains responses for each relevant issue, i.e., the answer candidates for each relevant issue. Subsequently, the \textit{Seeker} agent aggregates information from the responses, answer candidates, and the image caption, generating hypotheses with their confidence words, which construct a Multi-View Knowledge Base (MVKB). The \textit{Integrator} agent utilizes the MVKB from the \textit{Seeker} and the \textit{Responder} to get the voting pool and further votes to obtain the final answer. In this process, $H$ represents the Hypothesis. Given that there are 2 candidate answers, each relevant issue combined with the original question will generate four hypotheses, namely $H1$, $H2$, $H3$, and $H4$. To better illustrate how these Hypotheses are utilized in the interaction between the \textit{Integrator} and the \textit{Responder} agent, i.e., the step of \textbf{Re-answer Question}, we provide a single-step example in Figure~\ref{fig:one_step}.
}
  \label{fig:pipline}
\end{figure*}

In this section, we delineate the composition and functionalities of the three distinct agents, i.e., \textit{{Responder}}, {\textit{Seeker}} and {\textit{Integrator}, that constitute our SIRI. To effectively emulate the human-like top-down reasoning process for VQA tasks, our SIRI attempts to leverage the potential capability of VLMs by introducing the interaction dynamics among these three agents, as illustrated in Figure~\ref{fig:pipline}.
Initially, the \textit{Responder} agent $f_R$ generates a set of answer candidates, laying the groundwork for the subsequent reasoning process. Following this, the \textit{Seeker} agent $f_S$, acting as a central component of SIRI, generates relevant issues that can determine the answer from the provided answer candidates. Then, the \textit{Seeker} agent obtains the answer candidates for these relevant issues through the \textit{Responder} and acquires their corresponding hypotheses and their confidence scores via the LLM. Utilizing the capabilities of LLM, the \textit{Seeker} agent then compiles a Multi-View Knowledge Base (MVKB), which contains the correlations between the relevant issues and the question. Note that, these correlations are represented as hypotheses complete with confidence words assigned by the LLMs. Leveraging this MVKB, the \textit{Integrator} agent $f_I$ is equipped to select a refined answer from among the candidates. All prompts are shown in Figure~\ref{fig:prompts}. Overall, to solve the the given question-image pair $\left \langle Q,I\right \rangle$ to obtain the final answer $y^*$, our SIRI can be formalized as:
\begin{equation}
    y^* = SIRI(f_R, f_S, f_I| \textbf{Q}_{ac}), 
\end{equation}
where $\textbf{Q}_{ac}$ represent the answer candidates of the question $Q$ via $f_R$. 



\subsection{The Responder Agent}
Based on the VLM, the \textit{Responder} agent leverages the VLM to process the given question-image pair $\left \langle Q,I\right \rangle$ and generates the answer candidate set $\textbf{Q}_{ac}$ of $Q$ and the image caption $C$ of $I$ via two different prompts, i.e, $P_Q$ (e.g., Q + `` short question:'' as \cite{llava}) and $P_C$ (e.g., ``a photo of :'' as \cite{blip2}). Note that, $C$ is to provide the visual scene information to the \textit{Seeker} agent, thereby enhancing its ability to identify issues that are highly applicable to the current visual scene. Specifically, the \textit{Responder} agent $f_R$ based on the VLM $V$ with the fixed parameters $\theta$ directly generate the answer candidates $\textbf{Q}_{ac}$ and $C$ as follows:
\begin{equation}
\label{eq:R}
\begin{split}
    \textbf{Q}_{ac} &= f_R(P_Q,I,K; V_\theta)  \\   
    C &= f_R(P_C,I,\mathds{1}; V_\theta) \\
\end{split}
\end{equation}
where $K$ implies the size of the required answer candidate set.

The primary function of the \textit{Responder} agent is to answer the visual question directly through a simple methodology, acting as the foundational step in the top-down reasoning process. This initial response is critical, as subsequent processes within the framework hinge on the efficacy of this step. This is because the subsequent operation of the overall framework relies on the answer candidates provided in this step. Consequently, the number of answer candidates at this juncture is deemed a crucial parameter within our SIRI framework, i.e., the correct answer needs to be included among the top-K answer candidates.

Empirical evidence suggests that the correct answer is frequently found within the top-2 answer candidates. Statistics and analyses about the strategic emphasis on the top-2 candidate answers can be found in Section ~\ref{section:exper}. Based on this, our SIRI empirically sets the K value at 2. 


\subsection{The Seeker Agent}
The \textit{Seeker} agent $f_S$ is a crucial component of our SIRI, leveraging an LLM $L$ with the fixed parameters $\omega$ to find Relevant Issues $\mathbb{R}$ via prompt $P_R$ (as shown in Figure~\ref{fig:prompts} (a) ) and give their correlation with the question in the format of a hypothesis by the prompt $P_H$ (as shown in Figure~\ref{fig:prompts} (b) ), and its confidence score by $P_C$ (as shown in Figure~\ref{fig:prompts} (c) ). Firstly, it receives the question and answer candidates of the question from the \textit{Responder} agent, integrating them with the caption of the visual scene to generate relevant issues. It integrates the provided caption $C$ serving as the contextual scene information, to ensure the generated relevant issues are applicable to the current visual scene. This step can be formalized as follows:
\begin{equation}
\label{eq:S_ri}
\begin{split}
    \mathbb{R} = f_S(Q, \textbf{Q}_{ac}, C, P_R; L_\omega),
\end{split}
\end{equation}
The \textit{Seeker} agent then collaborates with the \textit{Responder} agent to obtain responses, i.e., answer candidates $\textbf{R}_{ac}=\{R_{ac}\}=\{(y, confidence)\}$ for the relevant issue, for these issues, crafting hypotheses with confidence scores. Note that, $y$ denotes the answer for relevant issue $r \in \mathbb{R}$. Each hypothesis combines an answer candidate $a_i \in \textbf{Q}_{ac}$ to the question with the answer candidate $\alpha_i \in \textbf{R}_{ac}$ to its relevant issue $r \in \mathbb{R}$. When the $K$ in top-$K$ (the number of answer candidates) is two, one relevant issue with the question can generate four hypotheses. It is critical to note that constructing hypotheses fundamentally depends on $\textbf{Q}_{ac}$ to the question $Q$. This dependency arises because the hypotheses are designed to elucidate the correlation between the relevant issues and the question, serving as a pivotal factor in determining which answer candidate is correct. We also use $f_S$ to obtain fluent hypotheses, as shown below:
\begin{equation}
\label{eq:S_h}
\begin{split}
    \textbf{H} = f_S(r, \textbf{R}_{ac}, Q, \textbf{Q}_{ac}, P_H; L_\omega),
\end{split}
\end{equation}
for each hypothesis $H \in \textbf{H}$, the \textit{Seeker} agent obtains its confidence score $O$ through $L_\omega$. The confidence score ranges from 0 to 1, with higher scores indicating a greater likelihood of the hypothesis occurring in reality. This simulates the probabilistic network of issue correlations that occurs when humans engage in top-down reasoning. When obtaining the $O$, the \textit{Seeker} agent also inputs the caption $C$ of the image, providing the LLM with visual scene information. It can be represented as follows:
\begin{equation}
\label{eq:S_cs}
\begin{split}
    \textbf{O} = f_S(C, \textbf{H}, P_C ; L_\omega).
\end{split}
\end{equation}
Providing scene information is important for delivering an accurate confidence score, as the confidence score is often strongly related to the specific scene. To illustrate this, consider the example where the question is `Is the dog a house dog?' and the hypothesis is `A dog without a tag is a house dog.' Initially, this hypothesis may receive a low confidence score because it is commonly believed that a wild dog would not have a tag. However, if the scene information reveals `A dog sitting on a bed.' the confidence score may be incorrect because a house dog inside a house may not necessarily wear a tag. On the other hand, a dog sitting on a bed is more likely to be a house dog. Therefore, without the scene information, the LLM may not provide an accurate confidence score, the same situation may occur in the step of obtaining relevant issues. Although LLMs can provide detailed confidence scores, the limited training data of VLMs makes it difficult for them to understand this expression form. Therefore, we convert the confidence score into a readable confidence word $\Phi$ via Eq.~(\ref{eq:confident_word}). For example, a confidence score of 0.2 corresponds to `Unlikely', while a score of 0.8 corresponds to `Likely'.

As previously highlighted, the \textit{Seeker} agent plays a critical role in our framework. The information it aggregates forms a useful resource for determining answer candidates. This repository of information termed the \textbf{M}ulti-\textbf{V}iew \textbf{K}nowledge \textbf{B}ase (MVKB), denoted as $M$, within the \textit{Seeker} agent, is depicted in Figure~\ref{fig:vision_kb}, illustrating its construction process. The MVKB encapsulates a series of hypotheses and their respective confidence word to the question, enriching the VLMs with not only additional visual clues but also insights into their applicability. This dual provision enables the VLMs to understand the relevance of this information, clarifying their contributory value in addressing the question. Furthermore, these confidence words provide the interpretability for our SIRI. The mechanics of the \textit{Seeker} agent are detailed in Algorithm \ref{alg:seeker}.

\begin{figure}[!t]
  \centering
  \includegraphics[width=0.9\linewidth]{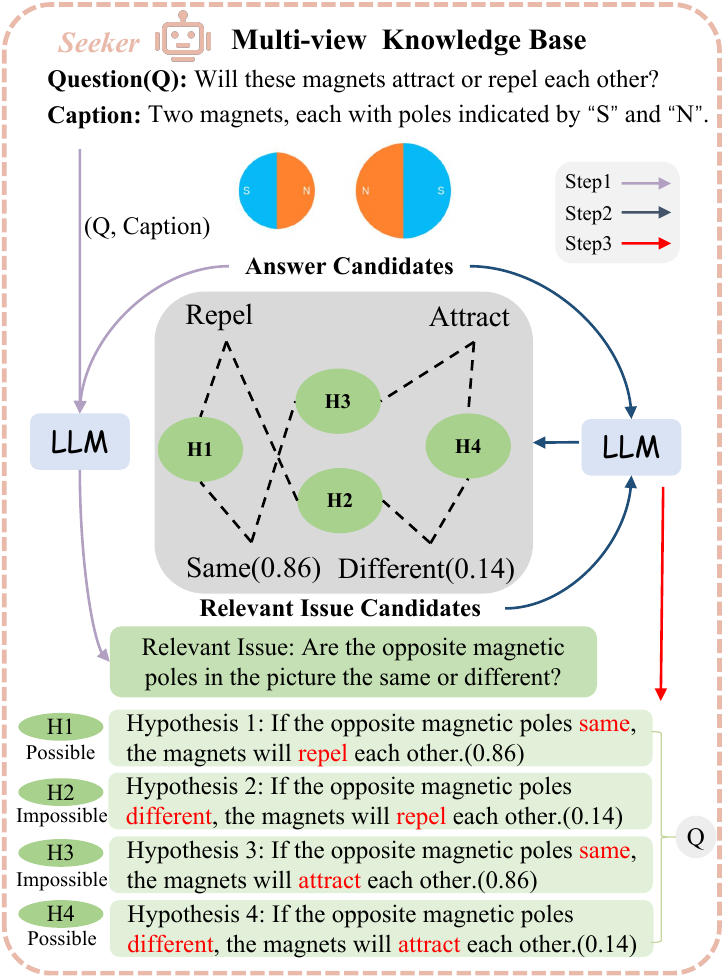}
  \caption{Multi-View Knowledge Base in \textit{Seeker}. The illustration highlights the distinctive characteristics of Multi-View Knowledge in response to different questions and the specific visual scene. The knowledge encompasses diverse hypotheses that establish connections with the question and the relevant issues in the given scene.
  }
    \label{fig:vision_kb}
\end{figure}

\begin{algorithm}[!t]
\caption{The Seeker Agent}
\begin{algorithmic}[1]
\small
\Procedure{Seeker}{$\textbf{Q}_{ac}$, $C$, $Q$} \Comment{$\textbf{Q}_{ac}$ is the answer candidates of the question given by the \textit{Responder} agent, $C$ is the visual scene caption also given by the \textit{Responder} agent via a prompt $P_C$.}

\State Obtain $\mathbb{R}$ via Eq.(\ref{eq:S_ri})
\For{$r$ in $\mathbb{R}$} 
    \State Obtain $\textbf{R}_{ac}$ via Eq.~(\ref{eq:R})
    \State Obtain $\textbf{H}$ via Eq. (\ref{eq:S_h}) \;
    \State Obtain $\textbf{O}$ via Eq. (\ref{eq:S_cs}) \; 
    \State $\Phi \leftarrow Convert(\textbf{O})$\; \Comment{Convert confidence score to confidence word, $\Phi$ means confidence word.}
    \State $M.append(\textbf{H}, \Phi, R_{ac} \in \textbf{R}_{ac}) $\; \Comment{$M$ means Multi-View Knowledge Base, $R_{ac}$ means the confidence of the hypothesis $\textbf{H}$ corresponding to the confidence score of relevant issue answer candidates $\textbf{R}_{ac}$. }
\EndFor
\State \textbf{return} $M$
\EndProcedure
\end{algorithmic}
\label{alg:seeker}
\end{algorithm}


\subsection{The Integrator Agent}
Upon the assembly of the Multi-View Knowledge Base, $M$, the \textit{Integrator} agent $f_I$ assumes the role of generating the final answer $y^*$, from the $\textbf{Q}_{ac}$ of the $Q$. It involves evaluating each $H$ along with its confidence word to the question through the $f_R$ to derive a refined answer. 

Despite the capabilities of VLMs, their interpretative limitations mean not every hypothesis necessarily yields a substantive impact on determining the final answer. In some instances, the derived answer may not align with the initial set of candidates. In response to this challenge and alignment with practices observed in other multi-agent system research\cite{scheuerman2021modeling,wang2023selfconsistency}, the \textit{Integrator} agent employs a voting mechanism. This approach not only capitalizes on the insights provided by the various hypotheses but also mitigates potential negative influences from less relevant or lower-quality $H$. To utilize the acquired Multi-View Knowledge Base in the \textit{Seeker} agent, i.e., to request the \textit{Responder} agent to obtain answers under different $H$ by concatenating the $H$, $\Phi$, and $Q$. To better illustrate how the Responder uses knowledge from the MVKB to provide new answers, we provide a single-step case in this step, as shown in Figure~\ref{fig:one_step}.

\begin{figure}[!t]
  \centering
  \includegraphics[width=\linewidth]{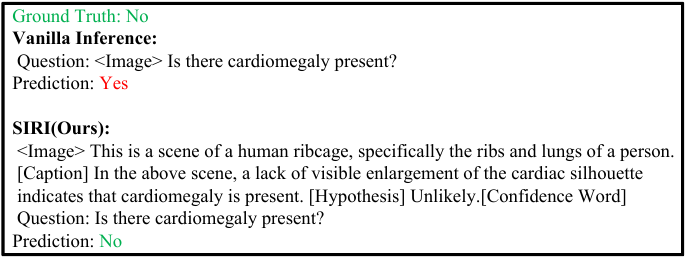}
  \caption{A single-step example where the \textit{Responder} agent is based on LLaVA-13B and the \textit{Seeker} agent is based on gpt4o-mini.}
    \label{fig:one_step}
\end{figure}

In the voting process of \textit{Integrator}, we employ a score-based weighted voting method. Each answer generated by a hypothesis is paired with the confidence of the corresponding answer candidate of the relevant issue, i.e., the $R_{ac}[confidence]$. This approach is used to merge the importance of different answer candidates of that relevant issue under the current image context. The process of the \textit{Integrator} agent is illustrated in Algorithm~\ref{alg:integrator}.

\begin{algorithm}[b]
\caption{The Integrator Agent}
\begin{algorithmic}[1]
\small
\Procedure{Integrator}{$M$, $I$, $Q$, $\textbf{Q}_{ac}$} \Comment{$M$ means the Multi-View Knowledge Base given by the \textit{Seeker} agent, $I$ is the given image, $Q$ is the corresponding question, and $\textbf{Q}_{ac}$ is the answer candidates of the $Q$.}

\For{$H$, $\Phi$, $R_{ac}$ in $M$} 
    \State $P_Q \leftarrow Concat(H,\Phi,I,Q)$\;
    \State $Q^* \leftarrow f_R(P_Q,I,\mathds{1};V_\theta)$\;\Comment{$Q^*$ the \textbf{top-1} answer from the response of the \textit{Responder} agent.}
    \If {$y$ in $\textbf{Q}_{ac}$}
        \State $Pool[y].append(R_{ac}[confidence])$  \Comment{Add the answer $y$ with its relevant issue confidence of $R_{ac}$ if the answer is in the answer candidates.}
    \EndIf
\EndFor
\State $y^* \leftarrow voting(Pool)$ \Comment{Vote the final answer $y^*$ with the highest score $\max_{y^*} Pool[y].sum()$ from $\textbf{Q}_{ac}$.}
\State \textbf{return} $y^*$
\EndProcedure
\end{algorithmic}
\label{alg:integrator}
\end{algorithm}

\begin{figure}[t]
  \centering
  \includegraphics[width=\linewidth]{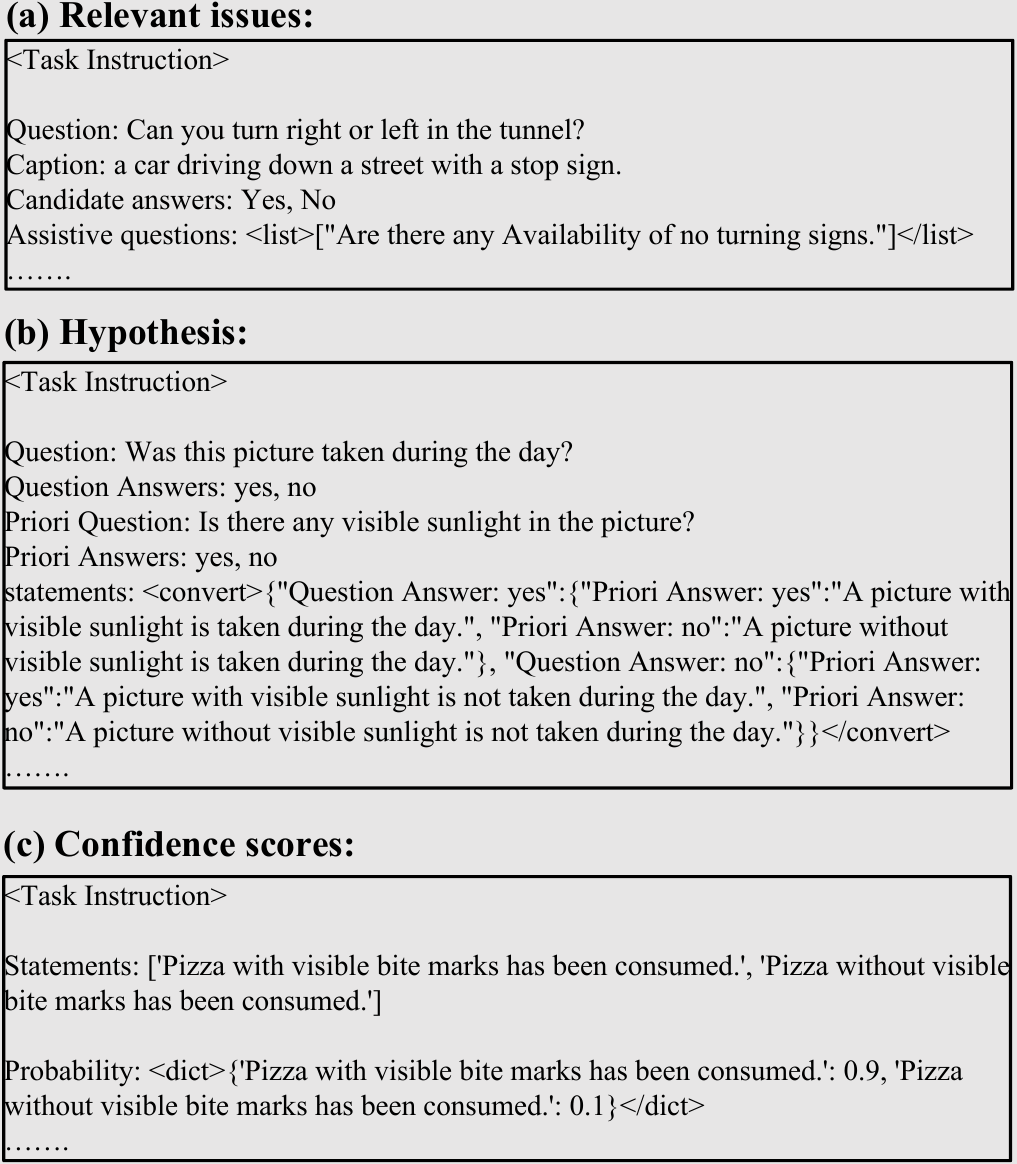}
  \caption{Prompt templates used in our SIRI for (a) Relevant Issues, (b) Hypothesis, and (c) Confidence Scores. Note that, we ignore the detailed task instructions and examples to highlight the content and structure of the prompts. 
  }
\label{fig:prompts}
\end{figure}
\section{Experiments}\label{section:exper}
\subsection{Experimental Settings}
\textbf{Datasets}. To validate the effectiveness of our SIRI, we conduct extensive evaluations on four diverse VQA datasets, each representing distinct domains and modalities: ScienceQA~\cite{scienceQA} (focusing on Science Knowledge Reasoning), A-OKVQA~\cite{AOKVQA} (emphasizing Common Sense and Outside Knowledge Answering), VQA-RAD~\cite{lau2018dataset} (dedicated to Medical Image Answering), and the Winoground~\cite{thrush2022winoground} (challenging Nature Image Reasoning). The details of the datasets are presented as follows:

\textbf{ScienceQA}~\cite{scienceQA} is derived from elementary and high school science curricula, offering a wide array of multimodal multiple-choice questions covering various scientific topics. We use the samples in the test set that have images.

\textbf{A-OKVQA}~\cite{AOKVQA} is a knowledge-intensive dataset requiring the integration of external knowledge for answering questions.  We use the val split in A-OKVQA.

\textbf{VQA-RAD}~\cite{lau2018dataset} comprises radiological images for medical image-based questioning. We use its test split for evaluation.

\textbf{Winoground}~\cite{thrush2022winoground} tests the capacity of the model for detailed understanding and visual reasoning, assessing its ability to accurately associate captions with images. Following ~\cite{khan2024exploring}, we convert Winoground to a VQA task with a boolean answer (yes or no) in the format of `does ``$\left \langle caption\right \rangle$" describe the image?'.


\textbf{Implementation Details.} In all experimental settings, we employ state-of-the-art VLMs across different model-sizes for the \textit{Responder} agent: BLIP-2 \cite{blip2} (Both ViT-g FlanT5XL 3B and ViT-g FlanT5XXL 11B) and LLaVA \cite{llava} (Both LLaVA-v1.5-7B and -13B). For the \textit{Seeker} agent, we use OpenAI's GPT-4o mini and GPT-3.5-turbo via the official API of OpenAI, served as our LLM, All experiments are conducted on a server with five RTX A6000 GPUs.

To \textbf{convert confidence score to confidence word}, we empirically use a five-tier to map the confidence score $O_i \in \textbf{O}$ into corresponding confidence word $\phi_i \in \Phi$. The categorization method is as follows:
\begin{equation}
\text{$\phi_i$} = 
\begin{cases} 
\text{Impossible} & \text{if } O_i < 0.2, \\
\text{Unlikely} & \text{if } 0.2 \leq O_i < 0.4, \\
\text{Possible} & \text{if } 0.4 \leq O_i < 0.7, \\
\text{Likely} & \text{if } 0.7 \leq O_i < 0.9, \\
\text{Probable} & \text{if } 0.9 \leq O_i \leq 1.
\end{cases}
\label{eq:confident_word}
\end{equation}
These mapping words are unoptimized, and exploring the relationship between the two to help VLMs better understand confidence scores remains a worthwhile endeavor.

As for \textbf{incorporating caption into hypothesis}, we use the following format to add the caption to the hypothesis: `This is a scene of ``$\left \langle caption \right \rangle$". In the above scene:  + Hyothesis'(The details format in the MVKB as shown in Figure~\ref{fig:pipline} and mentioned in Section~\ref{section:method}.)

\begin{table*}[htbp]
  \centering
    \caption{Zero-shot quantitative comparison with the SOTA VLM-based question decomposition method (Q-D)~\cite{khan2024exploring} on ScienceQA~\cite{scienceQA}, A-OKVQA~\cite{AOKVQA}, VQA-RAD~\cite{lau2018dataset} and Winoground~\cite{thrush2022winoground}. Our SIRI framework achieves superior performance than Q-D in thirteen out of sixteen comparisons, and a better average in all datasets. While enhancing performance, SIRI also provides interpretability for VQA tasks as shown in Figure \ref{fig:case_study}. \XSolidBrush indicates that the method does not achieve a performance improvement under this setting, and the \textbf{bold} indicates that under the same setting, the method achieves the best performance. Furthermore, we survey interpretability comparison among the baseline method, Q-D~\cite{khan2024exploring}, and our SIRI framework. The higher score for SIRI also validates its interpretability. Detailed results are shown in Table~\ref{tab:inter_tab}.}
\resizebox{2\columnwidth}{!}{
    \begin{tabular}{ccccccccccccc}
    \toprule
    Datasets & ScienceQA &       &       & A-OKVQA &       &       & VQA-RAD &       &       & Winoground &       &  \\
    \midrule
    Random Guess Acc & 28.70\% &       &       & 25\%  &       &       & None  &       &       & 50\%  &       &  \\
    \midrule
    LLM(Seeker) & \multicolumn{12}{c}{\textbf{GPT4o-mini}} \\
    \midrule
    VLM(Responder) &       & Acc   & $\eta$   &       & Acc   & $\eta$   &       & Acc   & $\eta$   &       & Acc   & $\eta$ \\
    \midrule
    \multirow{3}[0]{*}{BLIP2-FlanT5XL(3B)} & Baseline & 65.8\% & -     & Baseline & 64.4\% & -     & Baseline & 36.8\% & -     & Baseline & 55.2\% & - \\
          & Q-D   & 65.8\% & \XSolidBrush & Q-D   & 68.1(3.7)\% & 0.87  & \textbf{Q-D} & \textbf{38.9($\uparrow$2.1)\%} & 0.7   & Q-D   & 56.9($\uparrow$1.7)\% & 0.62 \\
          & \textbf{SIRI(Ours)} & \textbf{67.7($\uparrow$1.9)\%} & 0.52  & \textbf{SIRI(Ours)} & \textbf{69.0($\uparrow$4.6)\%} & 0.97  & SIRI(Ours) & 38.7($\uparrow$1.9)\% & 0.66  & \textbf{SIRI(Ours)} & \textbf{59.2($\uparrow$4.0)\%} & 0.98 \\
    \midrule
    \multirow{3}[1]{*}{BLIP2-FlanT5XXL(11B)} & Baseline & 64.3\% & -     & Baseline & 71.4\% & -     & Baseline & 56.5\% & -     & Baseline & 56.5\% & - \\
          & Q-D   & 64.8($\uparrow$0.5)\% & 0.77  & Q-D   & 71.8($\uparrow$0.4)\% & 0.58  & Q-D   & 61.2($\uparrow$4.7)\% & 0.62  & Q-D   & 56.5\% & \XSolidBrush \\
          & \textbf{SIRI(Ours)} & \textbf{70.3($\uparrow$6.0)\%} & 0.9   & \textbf{SIRI(Ours)} & \textbf{72.0($\uparrow$0.6)\%} & 0.65  & \textbf{SIRI(Ours)} & \textbf{62.2($\uparrow$5.7)\%} & 0.68  & \textbf{SIRI(Ours)} & \textbf{58.4($\uparrow$1.9)\%} & 0.77 \\
    \midrule
    \multirow{3}[2]{*}{LLaVA-v1.5(7B)} & Baseline & 66.8\% & -     & Baseline & 81.3\% & -     & Baseline & 37.4\% & -     & Baseline & 58.8\% & - \\
          & \textbf{Q-D} & \textbf{67.6($\uparrow$0.8)\%} & 0.56  & Q-D   & 81.3\% & \XSolidBrush & Q-D   & 38.9($\uparrow$1.5)\% & 0.57  & Q-D   & 58.8\% & \XSolidBrush \\
          & SIRI(Ours) & 67.2($\uparrow$0.4)\% & 0.87  & \textbf{SIRI(Ours)} & \textbf{81.4($\uparrow$0.1)\%} & 0.53  & \textbf{SIRI(Ours)} & \textbf{39.7($\uparrow$2.3)\%} & 0.93  & \textbf{SIRI(Ours)} & \textbf{61.1($\uparrow$2.3)\%} & 0.78 \\
    \midrule
    \multirow{3}[2]{*}{LLaVA-v1.5(13B)} & Baseline & 67.7\% & -     & Baseline & 82.6\% & -     & Baseline & 39.7\% & -     & Baseline & 58.7\% & - \\
          & Q-D   & 67.8($\uparrow$0.1)\% & 0.51  & \textbf{Q-D} & \textbf{83.6($\uparrow$1.0)\%} & 0.97  & Q-D   & 39.7\% & \XSolidBrush & Q-D   & 58.9($\uparrow$0.2)\% & 0.59 \\
          & \textbf{SIRI(Ours)} & \textbf{68.4($\uparrow$0.7)\%} & 0.69  & SIRI(Ours) & 83.5($\uparrow$0.9)\% & 0.99  & \textbf{SIRI(Ours)} & \textbf{45.8($\uparrow$6.1)\%} & 0.91  & \textbf{SIRI(Ours)} & \textbf{63.9($\uparrow$5.2)\%} & 0.84 \\
    \midrule
    \multirow{2}[2]{*}{Average} & Q-D   & \multicolumn{2}{c}{$\uparrow$0.35\%} & Q-D   & \multicolumn{2}{c}{$\uparrow$1.28\%} & Q-D   & \multicolumn{2}{c}{$\uparrow$2.10\%} & Q-D   & \multicolumn{2}{c}{$\uparrow$0.48\%} \\
          & \textbf{SIRI(Ours)} & \multicolumn{2}{c}{\textbf{$\uparrow$2.25\%}} & \textbf{SIRI(Ours)} & \multicolumn{2}{c}{\textbf{$\uparrow$1.55\%}} & \textbf{SIRI(Ours)} & \multicolumn{2}{c}{\textbf{$\uparrow$4\%}} & \textbf{SIRI(Ours)} & \multicolumn{2}{c}{\textbf{$\uparrow$3.35\%}} \\
    \bottomrule
    \end{tabular}%
    }
\label{tab:main_result_image}
\end{table*}%

\begin{table*}[htbp]

  \centering
    \caption{Comparison of the performance of our SIRI framework using GPT-3.5 and GPT4o-mini in the \textit{Seeker} agent. These results demonstrate that using more powerful LLMs can further improve the performance of our SIRI framework. It should be noted that due to a certain level of randomness in the output formats of different LLMs, there are slight differences in the baselines for GPT-3.5 and GPT4o-mini. Based on our manual review, the inferior format output of GPT-3.5, to some extent limits its effectiveness compared to 
    the more powerful LLM.}
  \resizebox{2\columnwidth}{!}{
    \begin{tabular}{ccccccccccc}
    \toprule
    Datasets & A-OKVQA &       &       & VQA-RAD &       &       & Winoground &       &       & Average \\
    \midrule
    LLM(Seeker) & \multicolumn{9}{c}{\textbf{GPT3.5-turbo}}                             &  \\
    \midrule
    VLM(Responder) &       & Acc   & $\eta$   &       & Acc   & $\eta$   &       & Acc   & $\eta$   &  \\
    \midrule
    \multirow{2}[2]{*}{BLIP2-FlanT5XL(3B)} & Baseline & 62.40\% & -     & Baseline & 36.50\% & -     & Baseline & 56.50\% & -     & \multirow{2}[2]{*}{$\uparrow$3.10\%} \\
          & SIRI(Ours) & \textbf{68.6($\uparrow$6.2)\%} & 0.96  & SIRI(Ours) & 37.8($\uparrow$1.3)\% & 0.64  & SIRI(Ours) & 58.2($\uparrow$1.7)\% & 0.87  &  \\
    \midrule
    \multirow{2}[2]{*}{LLaVA-v1.5(13B)} & Baseline & 82.60\% & -     & Baseline & 43.60\% & -     & Baseline & 58.50\% & -     & \multirow{2}[2]{*}{$\uparrow$1.80\%} \\
          & SIRI(Ours) & \textbf{82.8($\uparrow$0.9)\%} & 0.78  & SIRI(Ours) & 45.1($\uparrow$1.5)\% & 0.88  & SIRI(Ours) & 62.1($\uparrow$3.1)\% & 0.83  &  \\
    \midrule
    LLM(Seeker) & \multicolumn{9}{c}{\textbf{GPT4o-mini}}                               &  \\
    \midrule
    VLM(Responder) &       & Acc   & eta   &       & Acc   & eta   &       & Acc   & eta   &  \\
    \midrule
    \multirow{2}[1]{*}{BLIP2-FlanT5XL(3B)} & Baseline & 64.40\% & -     & Baseline & 36.80\% & -     & Baseline & 55.20\% &       & \multirow{2}[1]{*}{\textbf{$\uparrow$3.50\%}} \\
          & SIRI(Ours) & 69.0($\uparrow$4.6)\% & 0.97  & SIRI(Ours) & \textbf{38.7($\uparrow$1.9)\%} & 0.66  & SIRI(Ours) & \textbf{59.2($\uparrow$4.0)\%} & 0.98  &  \\
    \midrule
    \multirow{2}[1]{*}{LLaVA-v1.5(13B)} & Baseline & 82.60\% & -     & Baseline & 39.70\% & -     & Baseline & 58.70\% &       & \multirow{2}[1]{*}{\textbf{$\uparrow$4.10\%}} \\
          & SIRI(Ours) & \textbf{83.5($\uparrow$0.9)\%} & 0.99  & SIRI(Ours) & \textbf{45.8($\uparrow$6.1)\%} & 0.91  & SIRI(Ours) & \textbf{63.9($\uparrow$5.2)\%} & 0.84  &  \\
    \bottomrule
    \end{tabular}%
    }
  \label{tab:compare_llm}%
\end{table*}%

\begin{table*}[htbp]
\caption{Analysis of the quality of relevant issues. An $\tau$ value closer to 0.5 signifies that more relevant issues positively contribute to the correct answer. Additionally, filtering relevant issues can further enhance the performance of SIRI, it demonstrates that our SIRI framework has the potential for further performance improvement. For a fair comparison, we do not apply the optimized $\tau$ in Table \ref{tab:main_result_image}.}
\label{tab:quality_ri}
\resizebox{2\columnwidth}{!}{
  \centering
    \begin{tabular}{ccccccccccccc}
    \toprule
    Datasets & ScienceQA &       &       & A-OKVQA &       &       & VQA-RAD &       &       & Winoground &       &  \\
    \midrule
    Random Guess Acc & 28.7\% &       &       & 25\%  &       &       & None  &       &       & 50\%  &       &  \\
    \midrule
    LLM(Seeker) & \multicolumn{12}{c}{GPT4o-mini} \\
    \midrule
    VLM(Responder) &       & Acc   & $\tau$   &       & Acc   & $\tau$   &       & Acc   & $\tau$   &       & Acc   & $\tau$ \\
    \midrule
    \multirow{2}[2]{*}{BLIP2-FlanT5XL(3B)} & Baseline & 65.8\% & -     & Baseline & 64.4\% & -     & Baseline & 36.8\% & -     & Baseline & 55.2\% & - \\
          & SIRI(Ours) & 67.7($\uparrow$1.9)\% & 0.5   & SIRI(Ours) & 69.2($\uparrow$4.8)\% & 0.54  & SIRI(Ours) & 40.1($\uparrow$3.3)\% & 0.59  & SIRI(Ours) & 60.1($\uparrow$4.9)\% & 0.55 \\
    \midrule
    \multirow{2}[2]{*}{BLIP2-FlanT5XXL(11B)} & Baseline & 64.3\% & -     & Baseline & 71.4\% & -     & Baseline & 56.5\% & -     & Baseline & 56.5\% & - \\
          & SIRI(Ours) & 70.6($\uparrow$6.3)\% & 0.56  & SIRI(Ours) & 72.4($\uparrow$1.0)\% & 0.92  & SIRI(Ours) & 62.2($\uparrow$5.7)\% & 0.5   & SIRI(Ours) & 58.5($\uparrow$2.0)\% & 0.68 \\
    \midrule
    \multirow{2}[2]{*}{LLaVA-v1.5(7B)} & Baseline & 66.8\% & -     & Baseline & 81.3\% & -     & Baseline & 37.4\% & -     & Baseline & 58.8\% & - \\
          & SIRI(Ours) & 67.5($\uparrow$0.7)\% & 0.59  & SIRI(Ours) & 81.4($\uparrow$0.1)\% & 0.92  & SIRI(Ours) & 40.7($\uparrow$3.3)\% & 0.57  & SIRI(Ours) & 61.2($\uparrow$2.4)\% & 0.6 \\
    \midrule
    \multirow{2}[2]{*}{LLaVA-v1.5(13B)} & Baseline & 67.7\% & -     & Baseline & 82.6\% & -     & Baseline & 39.7\% & -     & Baseline & 58.7\% & - \\
          & SIRI(Ours) & 68.5($\uparrow$0.8)\% & 0.54  & SIRI(Ours) & 83.5($\uparrow$0.9)\% & 0.5   & SIRI(Ours) & 46.4($\uparrow$6.7)\% & 0.51  & SIRI(Ours) & 63.9($\uparrow$5.2)\% & 0.5 \\
    \midrule
    Average & SIRI(Ours) & \multicolumn{2}{c}{$\uparrow$2.43\%} & SIRI(Ours) & \multicolumn{2}{c}{$\uparrow$1.7\%} & SIRI(Ours) & \multicolumn{2}{c}{$\uparrow$4.8\%} & SIRI(Ours) & \multicolumn{2}{c}{$\uparrow$3.63\%} \\
    \bottomrule
    \end{tabular}%
    }

\end{table*}%

\begin{table*}[htbp]
  \centering
    \caption{The best performance of our SIRI framework with the \textbf{oracle} selection strategy. We use a posteriori-selection strategy to verify the best performance that SIRI can achieve. This indicates that with further improvement of the selection strategy, our SIRI framework has significant potential for further improvement.}
\resizebox{2\columnwidth}{!}{  
    \begin{tabular}{ccccccccc}
    \toprule
    Datasets & ScienceQA &       & A-OKVQA &       & VQA-RAD &       & Winoground &  \\
    \midrule
    LLM(Seeker) & \multicolumn{8}{c}{\textbf{GPT4o-mini}} \\
    \midrule
    VLM(Responder) &       & Acc   &       & Acc   &       & Acc   &       & Acc \\
    \midrule
    \multirow{2}[1]{*}{BLIP2-FlanT5XL(3B)} & Baseline & 65.80\% & Baseline & 64.40\% & Baseline & 36.80\% & Baseline & 55.20\% \\
          & SIRI\_Oracle(Ours) & 84.1($\uparrow$18.3)\% & SIRI\_Oracle(Ours) & 76.5($\uparrow$12.1)\% & SIRI\_Oracle(Ours) & 54.1($\uparrow$17.3)\% & SIRI\_Oracle(Ours) & 82($\uparrow$26.8)\% \\
    \midrule
    \multirow{2}[1]{*}{BLIP2-FlanT5XXL(11B)} & Baseline & 64.30\% & Baseline & 71.40\% & Baseline & 56.50\% & Baseline & 56.50\% \\
          & SIRI\_Oracle(Ours) & 77.7($\uparrow$13.4)\% & SIRI\_Oracle(Ours) & 73.7($\uparrow$2.3)\% & SIRI\_Oracle(Ours) & 90.0($\uparrow$33.5)\% & SIRI\_Oracle(Ours) & 88.9($\uparrow$32.4)\% \\
    \midrule
    \multirow{2}[2]{*}{LLaVA-v1.5(7B)} & Baseline & 66.80\% & Baseline & 81.30\% & Baseline & 37.40\% & Baseline & 58.80\% \\
          & SIRI(Ours) & 69.7($\uparrow$2.9)\% & SIRI\_Oracle(Ours) & 82.6($\uparrow$1.3)\% & SIRI\_Oracle(Ours) & 46.6($\uparrow$9.2)\% & SIRI\_Oracle(Ours) & 87.2($\uparrow$28.4)\% \\
    \midrule
    \multirow{2}[2]{*}{LLaVA-v1.5(13B)} & Baseline & 67.70\% & Baseline & 82.60\% & Baseline & 39.70\% & Baseline & 58.70\% \\
          & SIRI\_Oracle(Ours) & 69.5($\uparrow$1.8)\% & SIRI\_Oracle(Ours) & 84.7($\uparrow$2.1)\% & SIRI\_Oracle(Ours) & 54.2($\uparrow$14.5)\% & SIRI\_Oracle(Ours) & 84.5($\uparrow$25.8)\% \\
    \bottomrule
    \end{tabular}%
    }

  \label{tab:best_performence}%
\end{table*}%

\begin{figure*}[t]
  \centering
  \includegraphics[width=\linewidth]{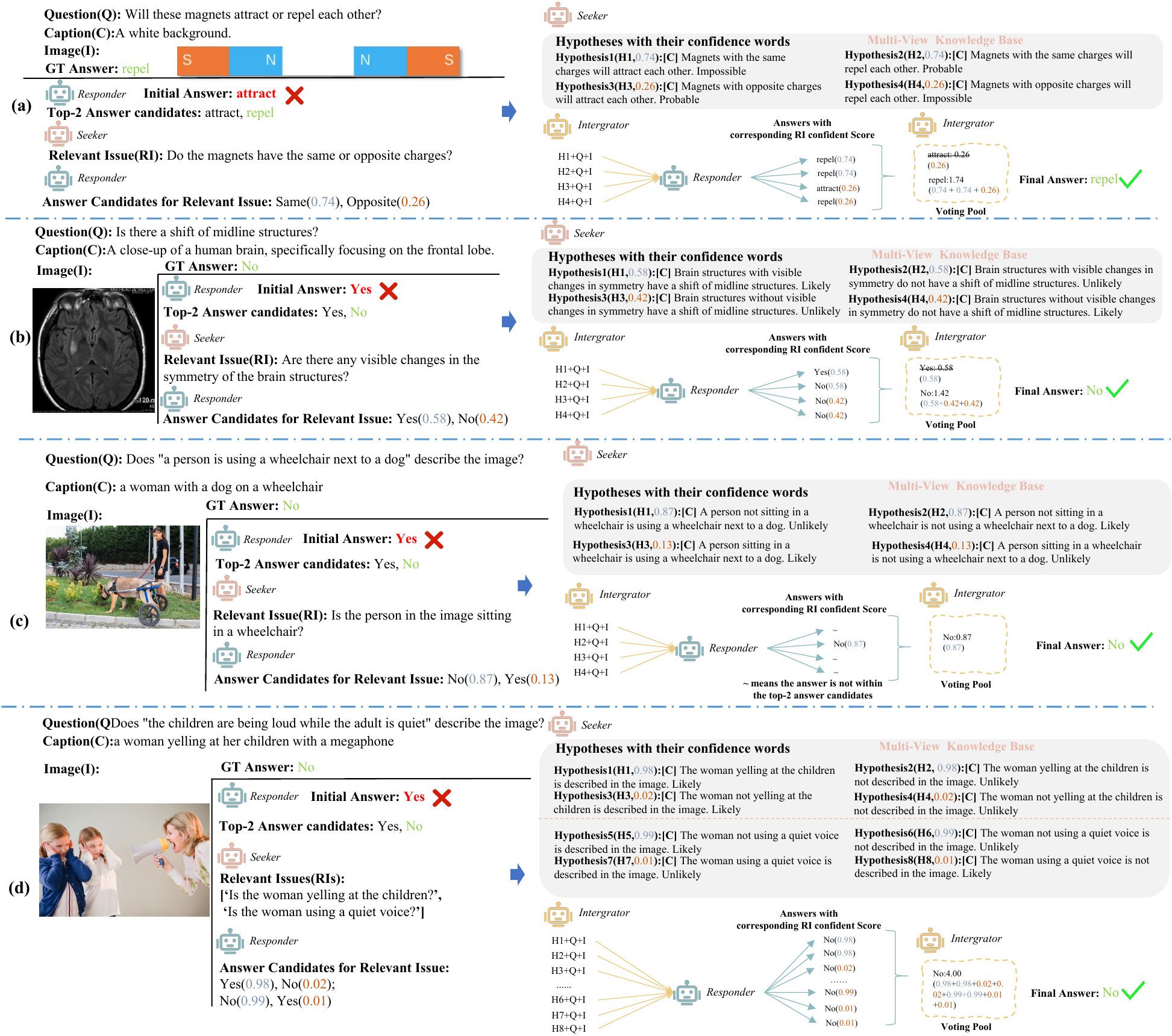}
  \caption{Examples from Winoground and VQA-RAD, which are incorrect originally but correct after the process of our SIRI ((a) is built with LLaVA-13B and gpt4o-mini, (b), (c) and (d) are built with LLaVA-7B with GPT3.5 turbo as the LLM). (a) shows SIRI uses the correlation between `Will these magnets attract or repel each other?' and `charges of the magnets'. (b) shows SIRI uses the visual scene information to find the correlation between the `a shift of midlines structures' and the brain.
  (c) shows a situation in which only one hypothesis generates an answer within the answer candidates, i.e., the H2. (d) shows that more than one relevant issue is used in the Seeker then generating a huge Multi-View Knowledge Base.}
\label{fig:case_study}
\end{figure*}

\subsection{Comparisons and Analyses}
\textbf{Baseline and Compared Methods}. We focus the comparison with the VLM-based methods on the VQA tasks. Specifically, we compare our SIRI with the single-step VQA method for VLM (Baseline), the current SOTA VLM-based question decomposition method, i.e., Q-D~\cite{khan2024exploring} ( the most relevant).


\textbf{Main Results.} We evaluate our SIRI on \textbf{zero-shot} VQA tasks and compare it with the other methods. As demonstrated in \cite{khan2024exploring}, applying such a `multi-step' approach to every question can have negative impacts. Similar to our comparison method, we introduce an extra hyperparameter $\eta$ to draw inspiration from human decision-making processes where answers with higher confidence scores are deemed more likely to be correct. For the confidence of the top-1 answer candidate higher than $\eta$, we apply its answer directly. For a fair comparison, we report the performance of all methods with their optimized $\eta$ thresholds with the same grid in Table~\ref{tab:main_result_image}. 

We provide the results for each comparison in Table \ref{tab:main_result_image} for the samples with at least one answer in the voting pool, which means the samples throughout SIRI. Our SIRI can improve the performance of the VLMs on VQA tasks on all baselines, the absolute improvement ranges from 0.1 to 6.1 points. Compared to the Q-D method, our SIRI achieves superior performance on most datasets and VLMs (thirteen of sixteen comparisons). In some experiments, Q-D is almost ineffective (e.g., BLIP2-FlanT5XL on ScienceQA, LLaVA-v1.5(13B) on VQA-RAD). Upon our manual inspection, as acknowledged in \cite{khan2024exploring}, it is very challenging for VLMs, which are smaller compared to LLMs, to decompose a question. This leads to nonsensical sub-questions in some cases. However, for SIRI, leveraging the powerful capabilities of LLMs can provide effective relevant issues even when facing difficult questions. Simultaneously, within our SIRI, the formulation of relevant issues is significantly aimed at distinguishing among answer candidates, which ensures that relevant issues are more closely related to the given question. Besides, we survey the interpretability of SIRI. SIRI receives higher interpretability scores compared to other methods, the results are shown in Table~\ref{tab:inter_tab}.

Our SIRI framework performs more effectively on the VQA-RAD and Winoground datasets, unlike the specialized datasets of ScienceQA and A-OKVQA, VQA-RAD and Winoground rely more on reasoning with common sense information and dependence on external knowledge. Unlike the specialized knowledge that is less distributed in the training set, providing correlation information for common sense can better assist VLMs in utilizing what they have already learned during training, thereby improving their performance. We also note that SIRI, with BLIP2-FlanT5XL(3B) as the VLM, shows outstanding performance on A-OKVQA, but the improvements are limited when the VLMs are BLIP2-FlanT5XXL(11B) and LLaVAs. This is because BLIP2-FlanT5XXL(11B) and LLaVAs already have a high baseline on A-OKVQA, making further improvements challenging.

To verify the sustainable improvement of our SIRI, we compare the effects when different LLMs serve as the \textit{Seeker} agent, the results are shown in Table~\ref{tab:compare_llm}. Compared to GPT-3.5, the more capable GPT4o-mini enables our SIRI to perform more outstandingly overall. This proves that with the enhancement of LLM, our SIRI obtains consistent performance improvements.






\subsection{More Analyses of Our SIRI}
\textbf{Quality of Relevant Issues}. Generally speaking, not all relevant issues may have a positive effect on our SIRI. To assess the quality of relevant issues, similar to $\eta$ for the question, we introduce a more hyperparameter $\tau$ for relevant issues. For the top-1 answer candidate confidence lower than $\tau$, our SIRI discards the relevant issue. As shown in Table~\ref{tab:quality_ri}, the optimized $\eta$ is close to 0.5 in most experiments, demonstrating that the majority of relevant issues positively influence the answer, indicating their high quality. Additionally, as the value of $\tau$ is empirically optimized for different settings, SIRI can achieve higher improvement points, further demonstrating the potential of our SIRI framework.

\textbf{The impact of $K$.} We conduct a statistical analysis to assess the upper limit of acc achievable by the top-2 answer candidates. Specifically, we examine the accuracy when the correct answer is within the top-2 answer candidates, rather than being limited to the top-1 answer candidate. This statistical evaluation serves to quantify the performance ceiling of our SIRI. As in Table \ref{tab:statistic_top_acc}, the top-2 setting achieves a high upper-limit performance for our SIRI. Thus, for a trade-off between the inference cost and accuracy, we empirically set $K$ as 2.


\begin{table}[]
\caption{Results of top-2 acc for BLIP2-FlanT5XL on VQA-RAD, A-OKVQA, Winoground and ScienceQA.}
\begin{tabular}{ccccc}
\hline
Dataset   & VQA-RAD & A-OKVQA & Winoground & ScienceQA \\ \hline
Top-1 Acc & 35.7\%  & 65.3\%  & 55.1\%     & 62.9\%    \\
Top-2 Acc & 64.7\%  & 83.8\%  & 99.9\%     & 76.9\%    \\ \hline
\end{tabular}
\label{tab:statistic_top_acc}
\end{table}

\textbf{The ceiling performance of our SIRI framework.} In both Table~\ref{tab:compare_methods} and Table~\ref{tab:quality_ri} settings, to ensure a fair evaluation and comparison, we apply the same filtering threshold to all questions or relevant issues. To further verify whether the top-down reasoning process can indeed help VLM provide a more accurate answer, we adopt an oracle post-hoc strategy, where each question selects the relevant issue that has the most positive impact on it. The experimental results are shown in Table~\ref{tab:best_performence}, demonstrating that the top-down reasoning strategy can significantly assist VLM in providing better answers.

\begin{table}[!t]
\caption{Ablation Study on \textit{Seeker}, and \textit{Integrator}. We conduct experiments with different strategies in the two agents}
\small
\centering
\begin{tabular}{lc}
\hline
Method                             & A-OKVQA     \\ \hline
SIRI(BLIP2 3B, GPT3.5)                               & 68.6\%      \\ \hline
\multicolumn{2}{c}{Ablation Study On Seeker}     \\ \hline
w/o answer candidates              & 66.8($\downarrow1.8$)\%    \\
w/o caption in confidence score    & 68.1($\downarrow0.5$)\%    \\
w/o caption in relevant issue      & 67.3($\downarrow1.3$)\%    \\
w/o converting confidence word      & 67.7($\downarrow0.9$)\%    \\ \hline
\multicolumn{2}{c}{Ablation Study on Integrator} \\ \hline
w/o confidence word                 & 66.8($\downarrow1.8$)\%    \\
w/o weighted voting                & 65.9($\downarrow2.7$)\%    \\
w/ relevant issue\&answer          & 65.9($\downarrow2.7$)\%   \\ \hline
\end{tabular}

\label{tab:ablation_study}
\end{table}

\textbf{Ablation Study.} We investigate the impact of different design strategies for \textit{Seeker} and \textit{Integrator} on overall performance as shown in Table \ref{tab:ablation_study} (Note that, the results for different VLM in the \textit{Responder} agent are been shown in Table \ref{tab:main_result_image}). For the \textit{Seeker} agent, we explore the effect of without the answer candidates of the question when generating relevant issues, without captions when obtaining relevant issues and confidence scores, and without converting the confidence score to the confidence word. The results show that removing answer candidates affects overall performance mostly. This is because, without answer candidates as prior information, relevant issues lack the specificity to differentiate between answer candidates. Moreover, the absence of scene information results in the relevant issues less applicable to the current scene. As for converting the confidence score to the confidence word, the results show that it helps VLMs better understand the correlation information in the MVKB. The effect of different types of LLM in the Seeker agent is shown in Table~\ref{tab:compare_llm}. For the \textit{Integrator} agent, we test its overall performance impact when not providing confidence word, and not implementing weighted voting. These two experiments demonstrate the effectiveness of correlation information and the efficacy of weighted voting in integrating the MVKB given by the \textit{Seeker} agent. We also explore the strategy that combines the relevant issues with its answer. Interestingly, this led to a significant drop in the result. We believe the impact it causes is error accumulation, which we detailed discuss in Section~\ref{section:disscussion}.

\textbf{Case Study.} We provide cases in Figure \ref{fig:case_study} to demonstrate how our SIRI improves the process of solving VQA tasks. In Figure~\ref{fig:case_study}(a), to answer the question, our SIRI generates hypotheses through the relevant issue of the state of the charges of the magnets. These hypotheses leverage the correlation knowledge that `magnets with the same/opposite charges will repel/attract each other.' to judge the `repel or attract' in the answer candidates of the question. Figure~\ref{fig:case_study}(b) presents a more complex scenario. To address the given question, SIRI uses the visual scene information provided by the caption to propose the relevant issue `Are there any visible changes in the symmetry of the brain structures?'. The hypotheses generated leverage the correlation knowledge that changes in symmetry likely indicate a shift of midline structures. In this example, without the visual scene information from the caption, our SIRI can not determine that the image is of the brain, let alone propose a relevant issue related to brain structures. Figure~\ref{fig:case_study}(c) and Figure~\ref{fig:case_study}(d) respectively depict scenarios where the regenerated answers are not all among the answer candidates and where there are multiple relevant issues. It is noteworthy that Figure~\ref{fig:case_study}(b) and Figure~\ref{fig:case_study}(d) show the capability to manage the accumulation of errors of our SIRI. For instance, in Figure~\ref{fig:case_study}(b), there is an erroneous answer for the relevant issue, and in Figure~\ref{fig:case_study}(d), the first relevant issue resulted in an incorrect confident word. More detailed discussions are presented in Section~\ref{section:disscussion}.

We also provide examples of the main types of failure cases in our SIRI, as shown in Figure~\ref{fig:failure_case}. In Figure~\ref{fig:failure_case}(a), although the \textit{Seeker} agent constructed a reasonable MVKB, the \textit{responder} agent provided an incorrect answer during the re-answer question step. Figure~\ref{fig:failure_case}(b) shows that the \textit{Seeker} agent issued a non-relevant issue that could not differentiate between answer candidates, while Figure~\ref{fig:failure_case}(c) demonstrates that a relevant issue could distinguish between answer candidates, but the \textit{Seeker} agent gives an inaccurate confidence score. To more clearly show the failed steps, we highlight the key part of the failure.

\begin{figure}[!b]
  \centering
  \includegraphics[width=0.95\linewidth]{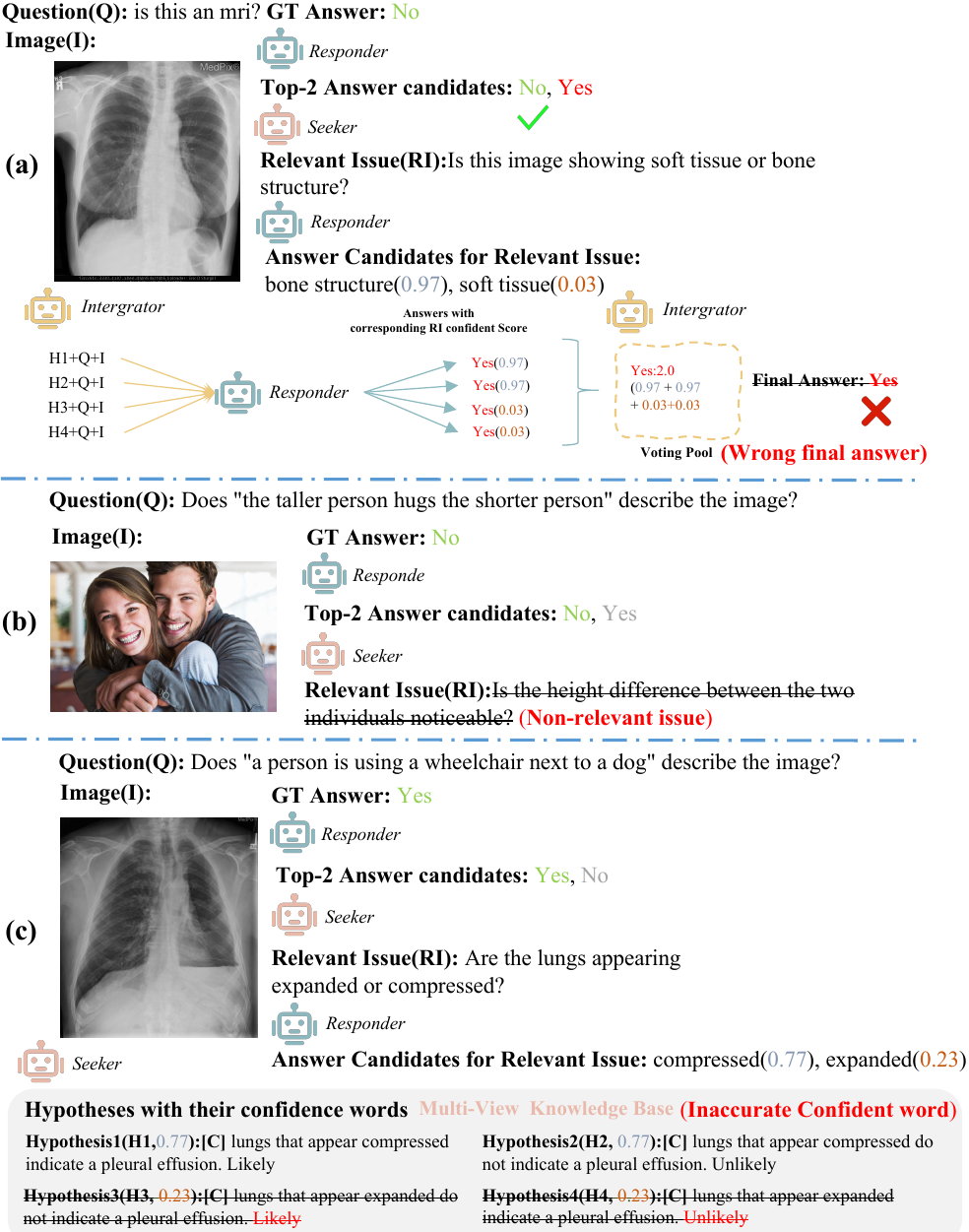}
  \caption{Failure examples of our SIRI. In (a), \textit{Responder} provides an incorrect answer during the re-answer process. In (b), the \textit{Seeker} agent presented a non-relevant issue that could not distinguish between the answer candidates to the question, and in (c), \textit{Seeker} provided an incorrect confidence word.
  }
    \label{fig:failure_case}
\end{figure}

\textbf{Interpretability.} Question-Decomposition (Q-D)~\cite{khan2024exploring} also provides an intermediate process for VQA tasks. To better compare the interpretability between SIRI and Q-D, we design a survey questionnaire for analysis. The designed survey consists of five questions, each offering three different methods in an anonymous form, i.e., our SIRI, Q-D~\cite{khan2024exploring}, and the baseline. Participants are required to rate each method on a scale from 0 to 5, where higher scores indicate better interpretability. The results are shown in Table~\ref{tab:inter_tab}.

For the single-step reasoning process of baseline, the lack of any displayed intermediate process results in a lack of interpretability in the answers of models. As for the Q-D, although it is also a multi-step reasoning approach, as pointed out in~\cite{khan2024exploring}, most decomposed questions are nonsensical due to the limited capabilities of VLMs. In contrast, our SIRI leveraging the powerful capabilities of LLMs, can effectively identify relevant issues related to the original question, ensuring not only effectiveness but also improved interpretability.

\begin{table}[]
\caption{Comparison of average interpretability scores for SIRI, Question-Decomposition, and the baseline(vanilla).}
\resizebox{1\columnwidth}{!}{
\centering
\begin{tabular}{cccc}
\hline
Method      & SIRI           & Question-Decomposition & Baseline \\ \hline
Ave. Rating & \textbf{4.114} & 2.258                  & 1.5      \\ \hline
\end{tabular}
}
\label{tab:inter_tab}
\end{table}

\section{Disscussion}\label{section:disscussion}
\textbf{Why SIRI now?} Reflecting on the constitute of our SIRI framework, it is worth examining the factors enabling the multi-agent collaboration framework. Firstly, the ability of the \textit{Responder} agent to address a wide of relevant issues marks a significant advancement over classification-based VQA models. Unlike models confined to classification tasks within a predefined answer set, the recent VLMs make \textit{Responder} extend to encompass open-world questions, thereby overcoming limitations associated with fixed responses. Secondly, the challenge of assigning accurate confidence scores to each hypothesis has been constrained by the reliance on static knowledge bases or dataset-specific statistics, both of which are prone to bias and limited by their scope. LLMs change this aspect by leveraging their extensive training corpora to provide unbiased confidence scores across a diverse range of hypotheses, which is like the insights in human top-down reasoning that derive from summarizing past experiences.

\textbf{How to avoid error accumulation in SIRI?} 
In the iterative interaction process of our SIRI framework, there is a natural propensity for errors to accumulate. To mitigate this, multiple strategies have been incorporated into the design of our SIRI. Initially, all intermediate outcomes are treated as provisional, with results being score-based; this approach ensures that no step within our SIRI takes any result as unequivocally accurate. Furthermore, to minimize the impact of incorrectly responded relevant issues on the overall effect, we only provide hypotheses and their confidence word derived from the \textit{Integrator} agent. Ablation studies have also proven the effectiveness of this strategy, as shown in Table \ref{tab:ablation_study}.

\textbf{Interpretability in SIRI.} The interpretability of our SIRI framework comes from two aspects. First, the explicit modeling of the correlation between relevant issues and the question, allows the VLM to understand how to utilize the provided information. The second is the score-based strategy throughout the framework, whether the confidence score or the final weighted voting, clearly demonstrates the decision-making process of our SIRI.

\textbf{Zero-Shot Setting.} The zero-shot setting is similar to the human-like top-down reasoning process. As mentioned in ~\ref{sec:intro}, humans can tackle VQA tasks by re-thinking from other perspectives rather than learning from a few examples. Thus, the zero-shot setting is more relevant to our SIRI, which aims at exploring the inherent potential of VLMs through the process of top-down reasoning.

\textbf{Relatiton to Chain-of-Thought (CoT)~\cite{wei2022chain}.} Generally speaking, our SIRI is a special cross-modal CoT mechanism and introduces the multi-agent framework to explicitly encourage the VLM and LLM to collaboratively solve the VQA task step-by-step. Specifically, our SIRI advances in emulating the human top-down reasoning process to evaluate the VQA answer candidates of VLM by proposing relevant issues via the relevant knowledge of LLM. This can effectively enhance the performance of
VLMs on VQA tasks.



\textbf{Limitation.} Although our SIRI framework shows effectiveness across various VQA datasets, the number of $K$ limits the potential for improving VQA task performance, meaning that the correct answer may not always be within the top-K candidates. Additionally, as $K$ increases, due to the need for more relevant issues to distinguish among these answer candidates, the overall computational and time costs of SIRI also rise. Furthermore, as the ceiling performance of our SIRI is shown in Figure~\ref{tab:best_performence}, how to further improve the filtering strategy to bring the performance of SIRI closer to its potential limit is also an important aspect that needs enhancement in the SIRI framework. Similarly, as shown in Table~\ref{tab:main_result_image}, the improvement effect of SIRI on datasets with a higher base accuracy is limited.

\section{Conclusion}\label{section:conclusion}

In this paper, we present SIRI, a novel multi-agent collaboration framework for VQA. Our SIRI leverages the knowledge of LLMs to facilitate a top-down reasoning process, towards mirroring human cognition. Comprising three distinct agents: \textit{Responder}, \textit{Seeker}, and \textit{Integrator}, our SIRI effectively seeks relevant issues and constructs the Multi-View Knowledge Base within a specific visual scene. Experimental results demonstrate the significant performance enhancement of our SIRI for various VLMs across various VQA datasets. In the future, we will extend our SIRI on the video-based VQA task by employing the existing video-based LLMs.





\ifCLASSOPTIONcaptionsoff
  \newpage
\fi

\bibliographystyle{unsrt}
\bibliography{egbib}

\newpage

\section{Biography Section}
\vspace{-13 mm}
\begin{IEEEbiography}[{\includegraphics[width=1in,height=1.25in, clip,keepaspectratio]{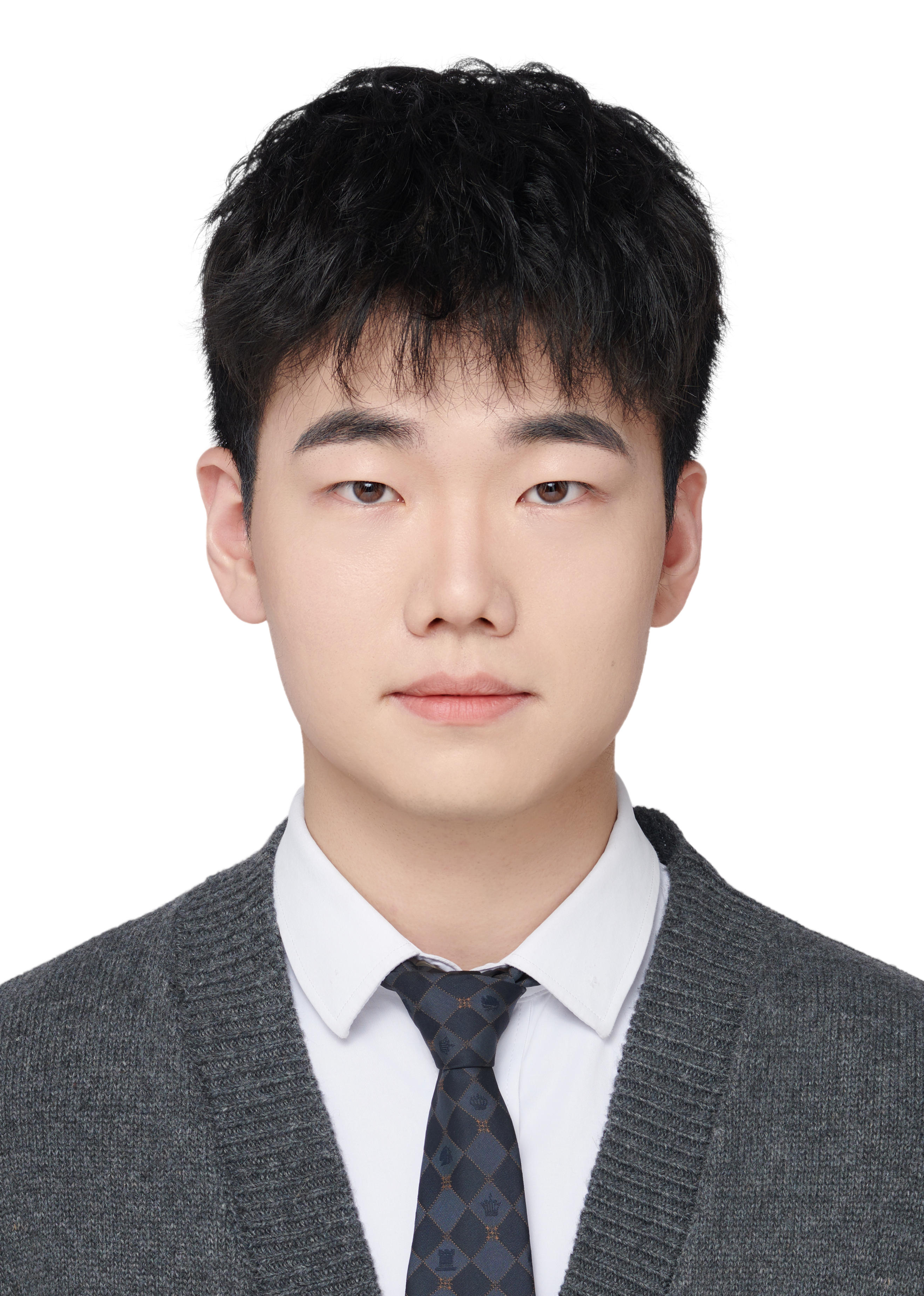}}]{Zeqing Wang} received the B.S. degree of software engineering from the College of Software, Jilin University, Changchun, China in 2023. He is currently a M.S. student at the School of Computer Science and Engineering, Sun Yat-sen University. His Main interests include multi-modal learning, multi-agent systems and vision language models. He has been serving as a reviewer for numerous academic journals and conferences, such as TMM, ACM MM.
\end{IEEEbiography}
\vspace{-15 mm}
\begin{IEEEbiography}[{\includegraphics[width=1in,height=1.25in, clip,keepaspectratio]{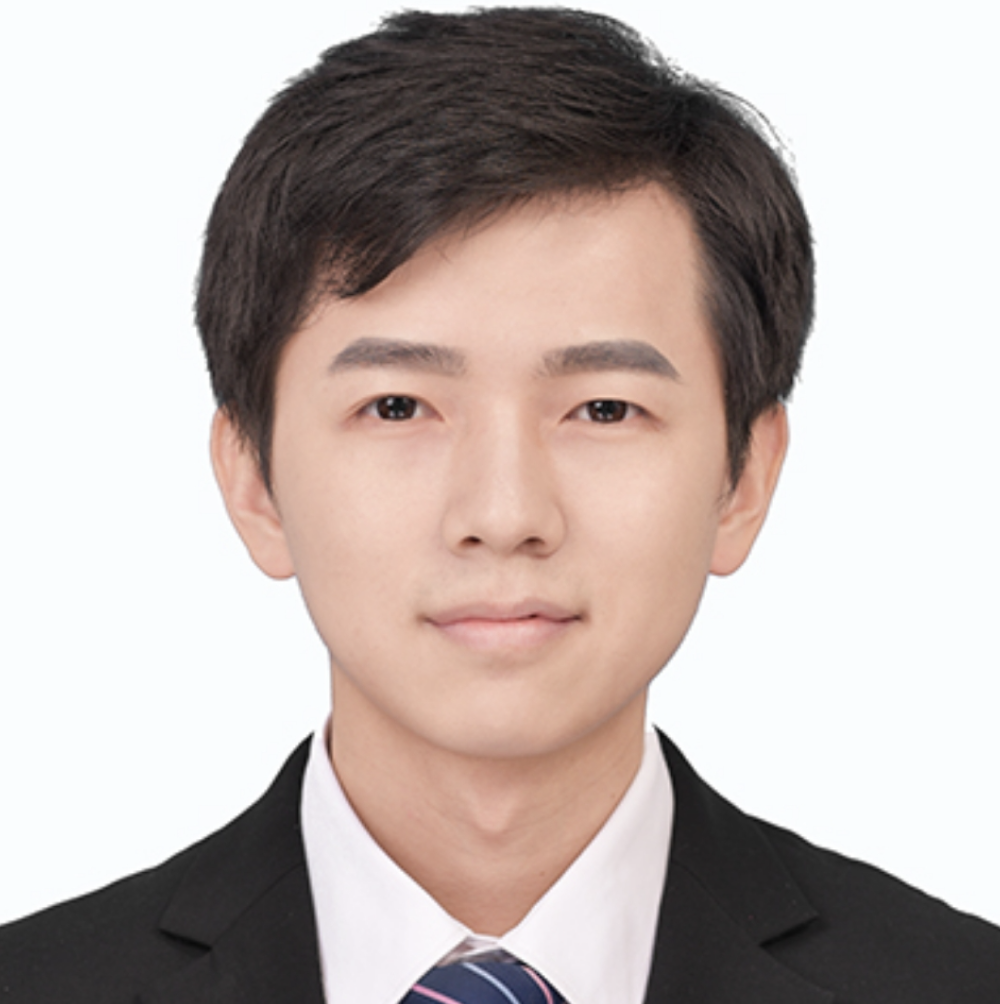}}]{Wentao Wan} is currently a Ph.D candidate at the School of Computer Science and Engineering, Sun Yat-sen University, since 2019. He received the M.S. degree in Computer Science and Technology from Huazhong University of Science and Technology in 2018, and the B.S. degree from Central South University in 2015. His research focuses on Neural-Symbolic systems, Visual Reasoning, and Multi-Modal Learning. 
\end{IEEEbiography}
\vspace{-14 mm}
\begin{IEEEbiography}[{\includegraphics[width=1in,height=1.25in, clip,keepaspectratio]{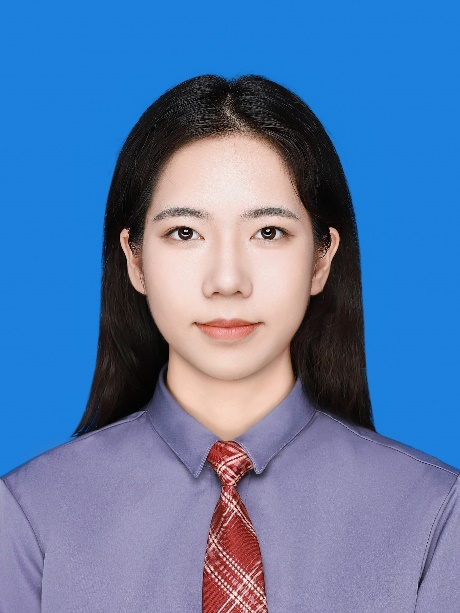}}]{Qiqing Lao} is currently a Master student at the School of Computer Science and Engineering, Sun Yat-sen University. She received the B.S. degree in Software Engineering from Sun Yat-sen University in 2024. Her research interests include machine learning, computer vision, and artificial intelligence. 
\end{IEEEbiography}
\vspace{-14 mm}
\begin{IEEEbiography}[{\includegraphics[width=1in,height=1.25in, clip,keepaspectratio]{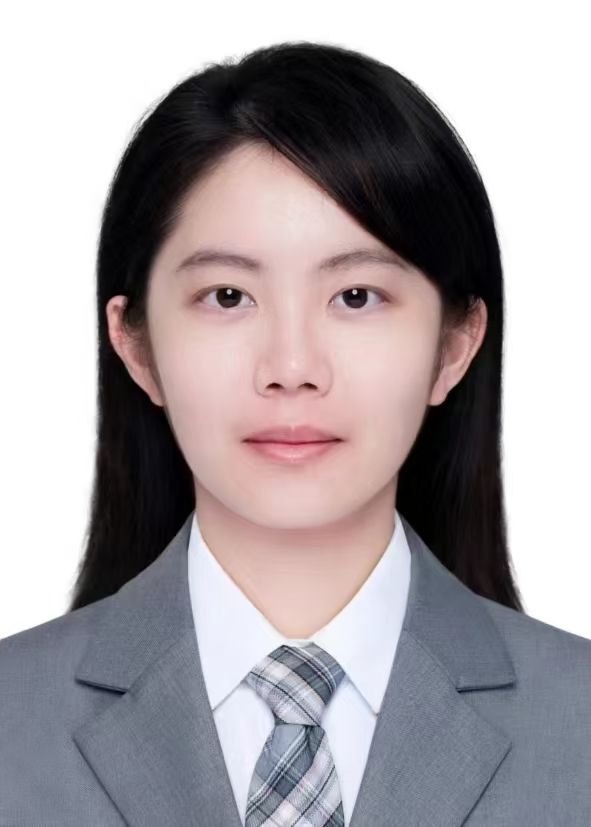}}]{Runmeng Chen} received her B.S degree in software engineering from the school of software, South China Normal University, Guangzhou, China in 2024. Her research interests include multi-agents systems, vision language model and down-stream tasks in vision-language models.
\end{IEEEbiography}
\vspace{-14 mm}
\begin{IEEEbiography}[{\includegraphics[width=1in,height=1.25in, clip,keepaspectratio]{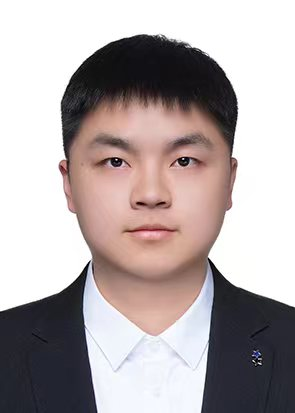}}]{Minjie Lang} received the Bachelor’s degree in Internet of Things Engineering from Northeastern University, China, in 2024, and joined Alibaba Group in the same year, marking the beginning of his professional career. He is currently serving as a Backend Development Engineer at Alibaba Group, where he is responsible for the e-commerce search engine. His work focuses on building efficient and stable system architectures to support massive data processing and user service demands.
\end{IEEEbiography}
\vspace{-13 mm}
\begin{IEEEbiography}[{\includegraphics[width=1in,height=1.25in, clip,keepaspectratio]{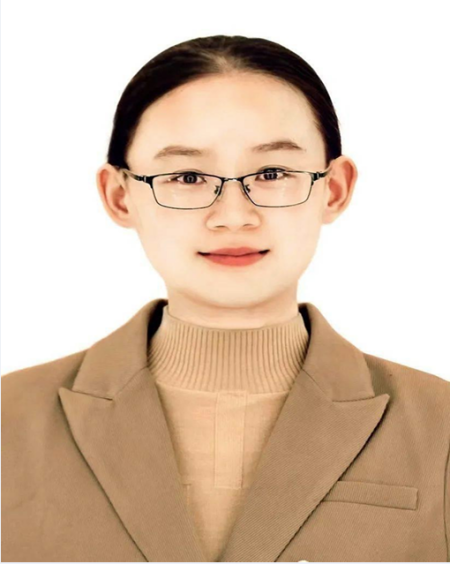}}]{Xiao Wang}(Senior Member, IEEE) received the Bachelor’s degree in network engineering from Dalian University of Technology, Dalian, China, in 2011, and the Ph.D. degree in social computing from the University of Chinese Academy of Sciences, Beijing, China, in 2016. She is currently a professor with the School of Artificial Intelligence, and Vice President of the Institute of Embodied Intelligence, Anhui University, Hefei, China. She is also Vice President of IEEE Intelligent Transportation Systems Society, and Chair of technical Committee 9.1 of IFAC. Her research interests include human behavior dynamics, social computing, embodied cognition, and AI Agent modeling. 
\end{IEEEbiography}
\vspace{-10 mm}
\begin{IEEEbiography}[{\includegraphics[width=1in,height=1.25in,clip,keepaspectratio]{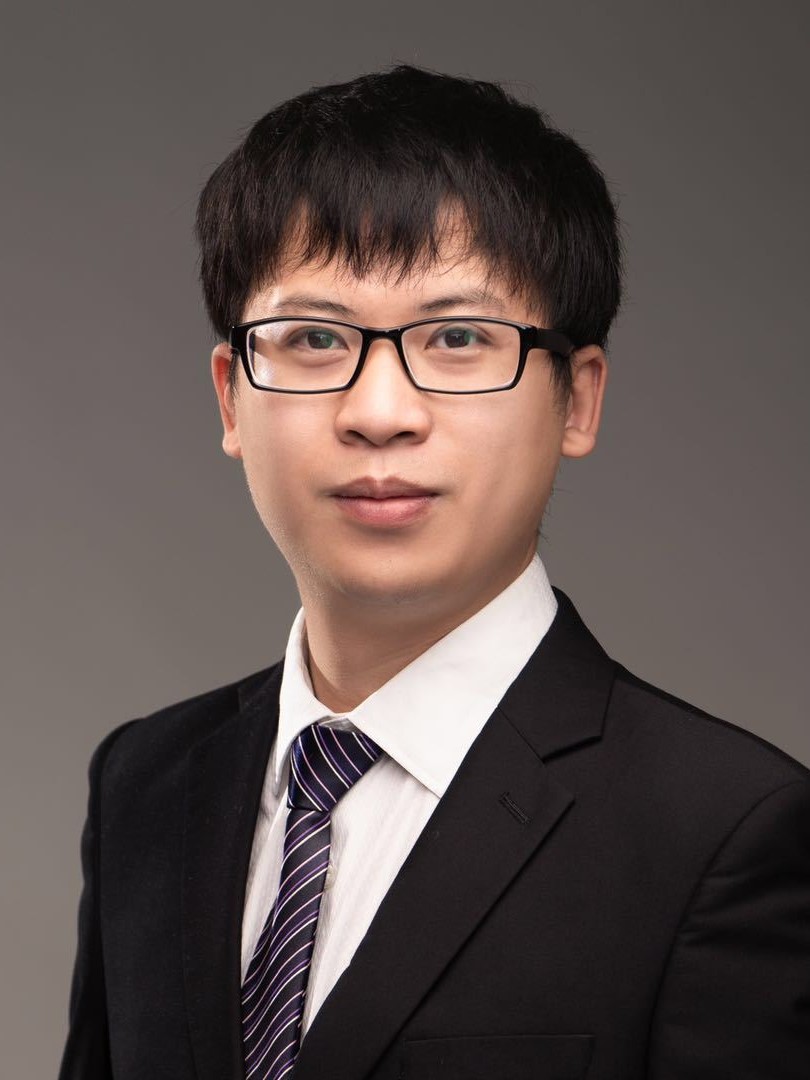}}]{Keze Wang} is nationally recognized as the Distinguished Young Scholars, currently serving as an Associate Professor at the School of Computer Science, Sun Yat-sen University, and a doctoral supervisor. He holds two Ph.D. degrees from Sun Yat-sen University (2017) and the Hong Kong Polytechnic University (2019). In 2018, he worked as a postdoctoral researcher at the University of California, Los Angeles, and returned to Sun Yat-sen University in 2021 as part of the ``Hundred Talents Program.'' Dr. Wang has focused on reducing deep learning's dependence on training samples and mining valuable information from massive unlabeled data.
\end{IEEEbiography}
\vspace{-125 mm}
\begin{IEEEbiography}
[{\includegraphics[width=1in,height=1.25in,clip,keepaspectratio]{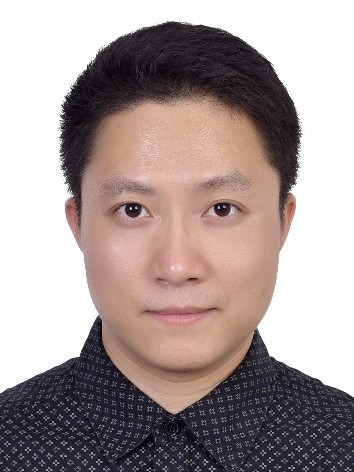}}]{Liang Lin}(IEEE Fellow) is a full professor of computer science with Sun Yat-sen University. He served as the executive director and distinguished scientist of SenseTime Group from 2016 to 2018, leading the R\&D teams for cutting-edge technology transferring. He has authored or co-authored more than 200 papers in leading academic journals and conferences, and his papers have been cited by more than 26\,000 times. He is an associate editor of \textit{IEEE Trans. Neural Networks and Learning Systems} and \textit{IEEE Trans. Multimedia}, and served as area chairs for numerous conferences, such as CVPR, ICCV, SIGKDD, and AAAI. He is the recipient of numerous awards and honors including Wu Wen-Jun Artificial Intelligence Award, the First Prize of China Society of Image and Graphics, ICCV Best Paper Nomination, in 2019, Annual Best Paper Award by Pattern Recognition (Elsevier), in 2018, Best Paper Dimond Award in IEEE ICME 2017, Google Faculty Award, in 2012. His supervised PhD students received ACM China Doctoral Dissertation Award, CCF Best Doctoral Dissertation and CAAI Best Doctoral Dissertation. He is a fellow of IET/IAPR.
\end{IEEEbiography}

\end{document}